\documentclass{article}

\usepackage{microtype}
\usepackage{graphicx}
\usepackage{subfigure}
\usepackage{booktabs} % for professional tables

\usepackage{url}
\usepackage{booktabs}       % professional-quality tables
\usepackage{nicefrac}       % compact symbols for 1/2, etc.
\usepackage{microtype}      % microtypography
\usepackage{xcolor}         % colors
\usepackage{graphicx}
\usepackage{wrapfig}
\usepackage{bm}
\usepackage{multirow}
\usepackage{soul}
\usepackage[font=footnotesize]{caption}
\usepackage{adjustbox}
\usepackage{multibib}
\usepackage{arydshln}
\usepackage{amssymb}
\usepackage{titletoc}
\usepackage{color, colortbl,booktabs}
\usepackage{comment}

\newcommand{\eg}{\emph{e.g.}}
\newcommand{\ie}{\emph{i.e.}}
\newcommand{\cf}{\emph{cf.}}

\newcommand{\wrt}{\emph{w.r.t.}}
\newcommand{\iid}{\emph{i.\thinspace{}i.\thinspace{}d.}}
\usepackage[accepted]{icml2023}

\usepackage[pagebackref]{hyperref}

\usepackage{xcolor}
\hypersetup{
  colorlinks   = true, %Colours links instead of ugly boxes
  urlcolor     = {green!50!black}, %Colour for external hyperlinks
  linkcolor    = {green!50!black}, %Colour of internal links
  citecolor   = {blue!50!black} %Colour of citations
}

\usepackage{amsmath}
\usepackage{amssymb}
\usepackage{mathtools}
\usepackage{amsthm}

\usepackage[capitalize,noabbrev]{cleveref}

\usepackage{glossaries}
\newacronym{mint}{MINT}{Maximum Information keypoiNTs}
\newacronym{poi}{PoI}{Points of Interest}
\newacronym{cv}{CV}{Computer Vision}
\newacronym{ml}{ML}{Machine Learning}
\newacronym{im}{IM}{information maximization}
\newacronym{mae}{ME}{masked entropy}
\newacronym{mce}{MCE}{masked conditional entropy}
\newacronym{it}{IT}{information transportation}
\newacronym{mi}{MI}{mutual information}
\newacronym{in}{IN}{Interaction Network}
\newacronym{dop}{DOP}{percentage of the detected object}
\newacronym{top}{TOP}{percentage of tracked objects}
\newacronym{rak}{RAK}{redundant keypoint assignment}
\newacronym{uak}{UAK}{unsuccessful keypoint assignment}
\newacronym{ise}{ISE}{image spatial entropy}
\newacronym{mme}{MME}{monkey model entropy
}
\makenoidxglossaries

\theoremstyle{plain}
\newtheorem{theorem}{Theorem}[section]
\newtheorem{proposition}[theorem]{Proposition}
\newtheorem{lemma}[theorem]{Lemma}
\newtheorem{corollary}[theorem]{Corollary}
\theoremstyle{definition}

\theoremstyle{remark}

\newcommand{\myparagraph}[1]{\textbf{#1}\hspace{0.2em}}

\usepackage[textsize=tiny]{todonotes}

\icmltitlerunning{Entropy-driven Unsupervised Keypoint Representation Learning in Videos}

\begin{document}

\twocolumn[
% \icmltitle{MINT: Unsupervised Keypoint Representation Learning in Videos with Information-Theoretic Measures}
\icmltitle{Entropy-driven Unsupervised Keypoint Representation Learning in Videos}




\begin{icmlauthorlist}
\icmlauthor{Ali Younes}{tu}
\icmlauthor{Simone Schaub-Meyer}{tu,hai}
\icmlauthor{Georgia Chalvatzaki}{tu,hai,cmbb}
% \icmlauthor{Firstname2 Lastname2}{equal,yyy,comp}
% \icmlauthor{Firstname3 Lastname3}{comp}
\end{icmlauthorlist}

\icmlaffiliation{tu}{Department of Computer Science, Technische Universität Darmstadt, Germany}
\icmlaffiliation{hai}{Hessian.AI}
\icmlaffiliation{cmbb}{Center for Mind, Brain and Behavior (CMBB), Uni. Marburg and JLU  Giessen}

\icmlcorrespondingauthor{Ali Younes}{ali.younes@tu-darmstadt.de}

\icmlkeywords{Representation Learning, Keypoint Discovery, Unsupervised Learning}

\vskip 0.3in
]



\printAffiliationsAndNotice{} % otherwise use the standard text.

%% begin text
\begin{abstract}
Extracting informative representations from videos is fundamental for effectively learning various downstream tasks. 
We present a novel approach for unsupervised learning of meaningful representations from videos, leveraging the concept of \glsfirst{ise} that quantifies the per-pixel information in an image. We argue that \textit{local entropy} of pixel neighborhoods and their temporal evolution create valuable intrinsic supervisory signals for learning prominent features.
Building on this idea, we abstract visual features into a concise representation of keypoints that act as \textit{dynamic information transmitters}, and design a deep learning model that learns, purely unsupervised, spatially \textit{and} temporally consistent representations \textit{directly} from video frames. Two original information-theoretic losses, computed from local entropy, guide our model to discover consistent keypoint representations; a loss that maximizes the spatial information covered by the keypoints and a loss that optimizes the keypoints' information transportation over time.
% , imposing consistency of information flow. 
We compare our keypoint representation to strong baselines for various downstream tasks, \eg, learning object dynamics.
Our empirical results show superior performance for our information-driven keypoints that resolve challenges like attendance to static and dynamic objects or objects abruptly entering and leaving the scene.
\footnote{ \url{https://sites.google.com/view/mint-kp}}
\end{abstract}

\section{Introduction}
Humans are remarkable for their ability to form representations of essential visual entities and store information to effectively learn downstream tasks from experience~\citep{cooper1990mental, radulescu2021human}. Research evidence shows that the human visual system processes visual information in two stages; first, it extracts sparse features of salient objects \citep{bruce2005saliency}; second, it discovers the interrelations of local features for grouping them to find correspondences~\citep{marr2010vision,kadir2001saliency}. For videos with dynamic entities, humans not only focus on dynamic objects, but also on the structure of the background scene if it plays a key role in the information flow~\citep{riche2012dynamic, human}. Ideally, we want a learning algorithm to extract similar sparse representations that can be useful for various downstream tasks. Notable research works in \gls{cv} and machine learning have proposed different feature representations from pixels
% for challenging downstream tasks
~\citep{szeliski,harris1988combined,lowe2004distinctive,rublee2011orb,mur2015orb}. In the deep learning era, convolutional neural network architectures have proven superior to handcrafted features, leading to new approaches for learning representations of \gls{poi} for tasks like localization and pose estimation~\citep{detone2018superpoint,ono2018lf,sarlin2019coarse,dusmanu2019d2,sarlin2020superglue}. 
% However, these methods usually extract an unstructured cloud of salient points that help to establish correspondences but lack interpretability. 

Keypoints stand out as sparse \gls{poi} \citep{jiang2009representations,alexe2010object} representing, \eg, objects~\citep{objectdetection}, human joints~\citep{pifpaf}, or structure useful for learning control~\citep{xiong2021learning}. Many keypoint detectors are trained in a supervised way~\citep{cao2017realtime}. Unsupervised and self-supervised learning can compensate the need for expensive human annotations \citep{ wang2020towards,videoprediction,control,gopalakrishnan2020unsupervised,scene}. Current state-of-the-art methods for unsupervised keypoint discovery focus mainly on dynamic entities in videos \citep{transporter,videostructre}. Namely, these methods are trained to reconstruct differences between frames, not effectively representing the scene's structure, while not easily disambiguating occlusions or consistently representing abruptly appearing/disappearing objects in a video.

We introduce \gls{mint}, an information-theoretic approach for unsupervised keypoint representation learning, treating keypoints as ``transmitters'' of prominent information in a video. Our proposed method relies on \textit{spatial information} computed in local neighborhoods (patches) around potential keypoints. We argue that the \gls{ise} \citep{brink1996using}, which quantifies the amount of local information of pixels in an image, and its evolution in a video, provide a strong \textit{inductive bias} for learning keypoint representations related to objects. Early works in \gls{cv} pointed out the relation of image entropy and object discovery~\citep{kadir2001saliency, bruce2005saliency, li2010visual}, but suffered from the need of filtering and tuning for every new setting to compute an accurate \gls{ise}~\citep{5200466}. Contrarily, our deep learning approach benefits from the approximation power of deep convolutional networks that learn nonlinear relations \textit{directly from image frames of a video} leading to spatially \textit{and} temporally consistent representations, that further generalize well. 
\begin{comment}
%, which are further efficient in inference, and generalize well to new settings.
%Using these classical concepts together with a neural network allows us to regularize to meaningful entities beyond low-level features unleashing the representation power to discover sparse spatio-temporal consistent features while keeping room for generalization.
%learn effectively consistent spatio-temporal representations of entities including the ability to dynamically adapt and generalize well to new scenes. 
%Leveraging these classical concepts within a learning framework allows our method to regularize to meaningful entities beyond the fixed low-level features unleashing the representation power of neural networks in order to discover sparse spatio-temporal consistent features, including the ability to generalize.
%while keeping the ability for generalization. % promoting generalization?
% The representation power of deep neural network allows our method to leverage these classical concepts to regularize to meaningful entities beyond low-level features unleashing the representation power to discover sparse spatio-temporal consistent features while promoting generalization.
%we convert these concepts into supervisory signals in form of our novel losses which can be used to learn representation of temporal consistent keypoints in an unsupervised deep learning framework.
%To compute \gls{ise}, we introduce a novel, efficient \textit{entropy layer} that operates locally on image patches. 

\end{comment}
\gls{mint} guides the spatio-temporal entropy coverage by the keypoints in a video, relying on an original formulation of unsupervised keypoint discovery with loss functions that \textit{maximize the represented image information entropy} and the \textit{information transportation across frames} by the keypoints, relying on a simple spatial entropy model and regularizers. Imposing spatio-temporal consistency of the represented entities
% , along with additional regularizers, 
enables \gls{mint} to effectively recover scene structure, allowing the subsequent \textit{simultaneous} detection and tracking of objects.

We provide qualitative and quantitative empirical results on four different video-datasets against strong baselines for unsupervised temporal keypoint discovery% \footnote{\url{https://sites.google.com/view/mint-kp}}
, unveiling the superior representation power of \gls{mint}. %Unsupervised keypoint discovery is challenging to evaluate due to the absence of designated metrics and benchmarks. Here, 
To address the challenge of quantitative evaluation of unsupervised keypoint discovery due to the absence of designated datasets, we provide a set of new metrics and a benchmark based on videos from CLEVRER~\citep{clevrer}. Moreover, we provide results on two challenging datasets, MIME \citep{mime} and  SIMITATE \citep{simitate}, that contain realistic scenes of various difficulties (close-up frames with dynamic interactions vs. high-res wide frames with clutter). We show that \gls{mint} economizes the use of keypoints, deactivating excessive ones when the information is well contained, and dynamically activating them to represent new entities entering the scene temporarily. Finally, to demonstrate the suitability of \gls{mint} as a representation for control, we devise an imitation learning downstream task on environments from MAGICAL~\citep{magical}.

\textbf{Contributions.} In summary, we introduce: 
\emph{(1)} an original unsupervised keypoint representation learning approach using information-theoretic measures, via the classical concept of \gls{ise} that inspired us to postulate keypoints as information transmitters;
\emph{(2)} an entropy layer for computing spatial image entropy 
% for patches 
efficiently; %, which we make public; 
\emph{(3)} an unsupervised way for representing variable number of entities in videos by switching on/off keypoints for covering spatio-temporal information; and
\emph{(4)} a new set of evaluation metrics for an intuitive downstream task for benchmarking the performance of unsupervised temporal keypoint discovery methods.

\section{Maximum Information Keypoints}\label{sec:method}
We propose an unsupervised method for keypoint discovery in videos based on information-theoretic principles. Keypoints should adequately represent the scene and the dynamic changes in it. Starting from our original assumption that a keypoint represents the spatial information of a patch of an image frame, we leverage the classical concept of \gls{ise}~\citep{brink1996using,razlighi2009comparison} to measure the amount of information represented by a keypoint. We argue that keypoints should cover areas in the image that are rich in information, while the number of keypoints should dynamically adapt to represent new information. Finally, keypoints should consistently represent the same information pattern spatio-temporally in a video. With this motivation, we propose to maximize the information covered by the keypoint representation in a video by introducing original losses for unsupervised temporal keypoint discovery. We mainly introduce two losses based on information-theoretic measures:
\emph{(1)} An \textit{information maximization loss} that encourages the keypoints to cover areas with high spatial entropy in a single frame. \emph{(2)} An \textit{information transportation loss} that enables the keypoints to represent the same entity over subsequent frames. 
We present these losses and theoretical analyses supporting their design in the following.
\begin{figure*}[ht!]
    \centering
    \includegraphics[width=0.85\linewidth]
    {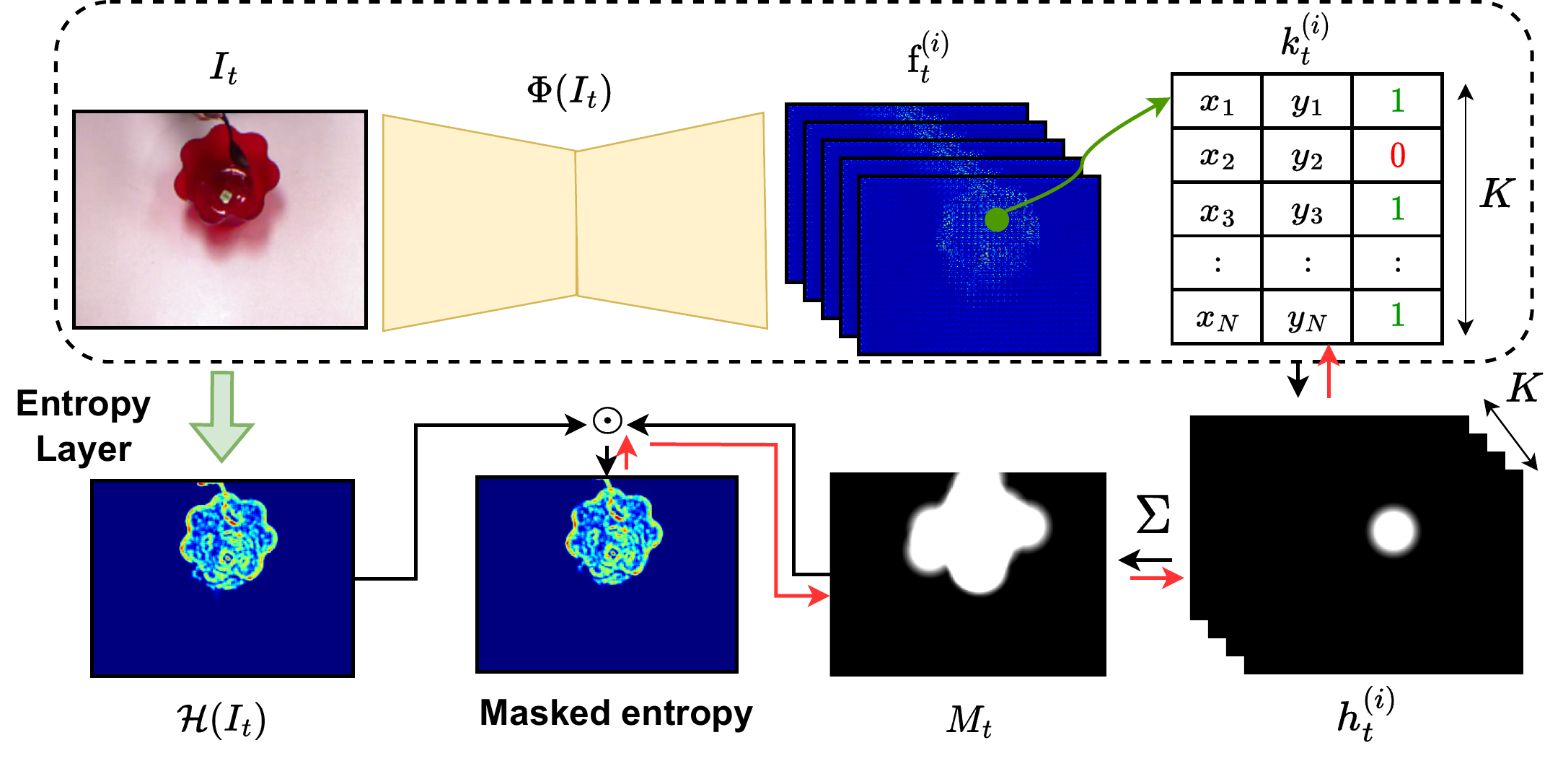}
    \vspace{-0.6cm}
    \caption{The architecture of our keypoint model $\Phi(I_t)$ (\cref{sec:arch}) and the masked entropy (\cref{sec:infomax}). For an input image $I_t$ our model $\Phi(I_t)$ outputs $K$ feature maps $\text{f}_t^{(i)}$ for each keypoint $\text{k}_t^{(i)}\text{, } i \in \{1, \dots, K\}$. A heatmap $h_t^{(i)}$ is generated for each keypoint, while the active keypoints are aggregated to form the mask $M_t$. The entropy layer computes the entropy of the image $\mathcal{H}(I_t)$. Our \gls{mae} loss maximizes the percentage of the entropy in the masked entropy image. Red arrows show the backward gradient flow. Only the part encircled by the dashed line is used during inference.}
    \label{fig:infomax}
    \vspace{-0.5cm}
\end{figure*}
\subsection{Image Spatial Entropy (ISE)}\label{sec:entropy}
%Images render the world into a set of pixel values lying on a discrete two-dimensional grid. Our information-theoretic treatment starts by quantifying the amount of information at each pixel location in a single image frame. 
Our information-theoretic approach for unsupervised keypoint discovery requires quantifying the amount of information each pixel location in a single frame carries. 
%Our original approach considers classical ideas in~\gls{cv}, regarding object detection from saliency. 
We leverage the idea of computing the information of patches in an image (local neighborhoods around a keypoint), using the classical concept of \gls{ise}~\citep{razlighi2009comparison}. \gls{ise} provides the pixel-wise information in the spatial domain of the image, and it has been greatly explored in computer vision, \eg, in Markov Random Fields ~\citep{5200466}. 
Images can be considered as lattices where pixels are random variables~\citep{li2009markov}.
We compute the discrete probability density of a pixel using the statistics of the color intensities in its neighborhood, represented by a normalized histogram of the neighboring pixel values \citep{sabuncu2006entropy}. This way of computing \gls{ise}~\citep{razlighi2009comparison} assumes that pixels in the image lattice are \iid and their entropy is computed using Shannon's definition~\citep{shannon2001mathematical} based on the probability of each pixel. 
%We use the statistics of the intensities of the neighboring pixels to compute the discrete probability density %in the neighborhood of a pixel using the normalized histogram of the pixel values \citep{thesisentropy} in the neighborhood. 
To compute these histograms efficiently and to derive the final \gls{ise}, we developed a computationally optimized entropy layer as detailed in \cref{sec:a_entropy}.

Our entropy layer estimates the pixel-wise image spatial entropy \gls{ise} $\mathcal{H}(I)$ for an RGB input image $I \in \mathbb{R}^{\mathrm{H} \times \mathrm{W} \times 3}$, with $\mathrm{H}$ being the height and $\mathrm{W}$ the width of an image frame with $3$ color channels. $\mathcal{H}(I)$ consists of the local entropies $\mathcal{H}(I(x,y))$ computed at each pixel location $(x,y)$ by estimating the entropy of the neighborhood region $R(x,y)$ centered at $(x,y)$, using a normalized histogram-based discrete probability function $p(b,R(x,y))$ for each color value $b$ in the region $R(x,y)$ summed and normalized over the color channels (details in \cref{sec:a_entropy}). The final per-pixel local entropy is
 \begin{equation}\mathcal{H}(I(x,y))= - \sum_{b=0}^{255} p(b,R(x,y)) \log(p(b,R(x,y))).
\end{equation}
% Computing the local entropy for all pixels results in the image spatial entropy %\gls{ise} 
% $\mathcal{H}(I)$.% (the use of image refers to the 2-dimensional representation of the local entropies on the image lattice).
% The following sections present a way to use the local image entropy $E(I)$ with information-theoretic principles for keypoint discovery in videos.

% \vspace{-0.5cm}
\subsection{Entropy-driven Keypoint Discovery}
\label{sec:arch}
We consider keypoints as a compact sparse representation of images, which attend to prominent entities in a scene \citep{szeliski}. Keypoints should represent distinctive information patterns overlaid on a set of neighboring pixels (patches) in an image frame. We explicitly treat the keypoint (at the center of a patch) as the information transmitter of its neighborhood. Based on \gls{ise}~\citep{razlighi2009comparison}, we compute the spatial entropy of each keypoint, which allows for developing an end-to-end unsupervised keypoint discovery approach using information-theoretic measures. Maximizing the keypoint information acts as an \textit{intrinsic inductive bias} for learning to represent areas of high entropy. Although a simple model to compute \gls{ise} can lead to local entropy overestimation~\citep{brink1996using}, we show empirically (cf.~\cref{sec:exper}) that when we regularize the proposed losses effectively, we get useful, well-behaved keypoint representations.
%very good performance, suppressing overestimation bias.

We define a keypoint discovery model $\Phi ({I}_t)$ (cf.~\cref{fig:infomax}), which is a deep neural network that discovers $K$ keypoints $\text{k}_t^{(i)}$, $i \in \{1,...,K\}$, in an input color image ${I}_t$ at time $t$. %\in \mathbb{R}^{H\times W \times 3 }$ of height $H$ and width $W$. 
It outputs $K$ feature maps $\text{f}_t^{(i)}$, each corresponding to one keypoint. The coordinates $(x_i,y_i)_t$ of the respective keypoint $k_t^{(i)}$ are obtained with a spatial soft-argmax~\citep{softargmax}. Besides predicting the coordinates, the model also assigns an activation status $s_t^{(i)}=\{0,1\}$ per keypoint. The activation status determines whether a keypoint is active ($s_t^{(i)} = 1$) or not ($s_t^{(i)}=0$) in a specific frame $t$, allowing the network to decide on the ideal number of active keypoints. Overall, a keypoint is defined by its coordinates and the activation score $\text{k}_t^{(i)} = (x_i,y_i,s^{(i)})_t$. 
To get the information coverage, we define a differentiable heatmap $h^{(i)}_t\in \mathbb{R}^{H\times W}$ for each $i^{\text{th}}$ keypoint by thresholding a distance-based Gaussian $G^{(i)}_t$ centered at the coordinates of the keypoint (details in \cref{sec:heatmap}).
% Hence, The heatmaps localize the keypoint information coverage. 
As we want to maximize information coverage by the keypoints spatio-temporally, we need to ensure that both the inter-frame and intra-frame information is sufficiently transmitted. Inspired by information theory, we derive novel losses that allow us to learn information-driven keypoint representations while providing error bounds that theoretically justify the design of those losses~\citep{sabuncu2006entropy,yu2021information}.

\subsubsection{Maximizing Keypoint Information} \label{sec:infomax}
With information maximization, we encourage keypoints to represent image regions rich in information (high spatial entropy). We want to enforce maximum collective spatial information coverage by the keypoints for representing all entities in a frame. For that, we introduce two losses: the \glsfirst{mae} loss and the \gls{mce} loss.

\begin{figure}[t!]
    \centering
    \includegraphics[width=\linewidth]{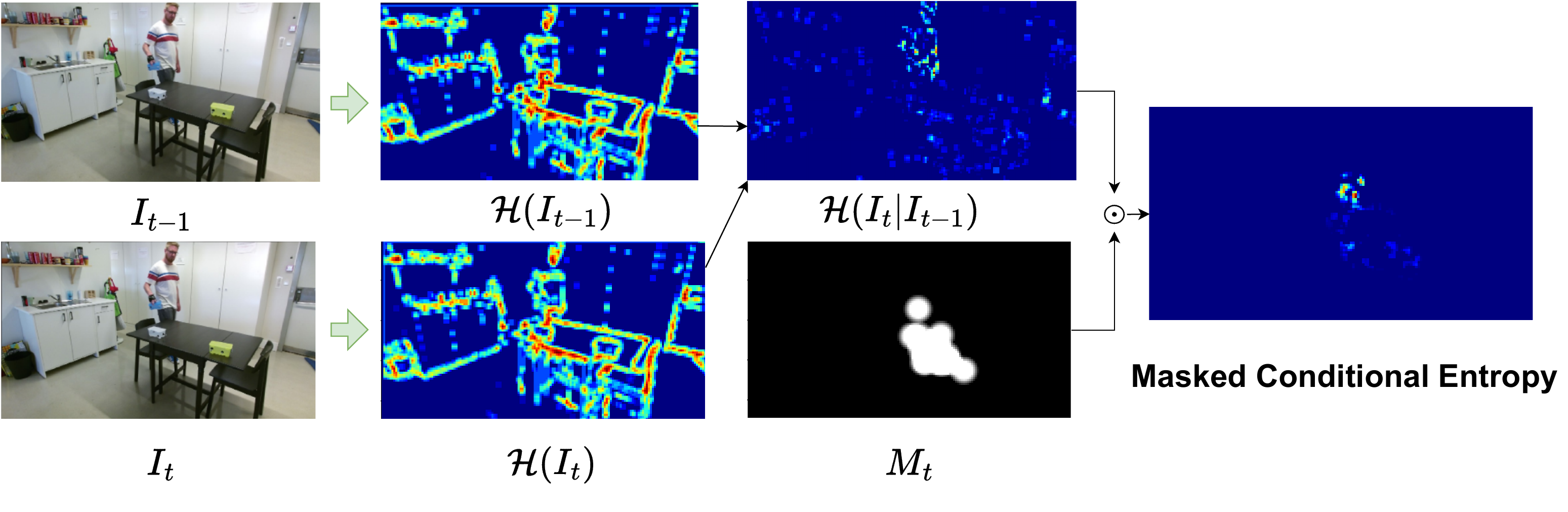}
    \vspace{-0.6cm}
    \caption{Masked conditional entropy (MCE) computation. Given two consecutive images $I_{t-1}$ and $I_t$, we extract their spatial entropies $\mathcal{H}(I_{t-1})$ and $\mathcal{H}(I_t)$. The conditional spatial entropy $\mathcal{H}(I_{t}| I_{t-1})$ depends on the spatial entropy of both images. Multiplying the conditional spatial entropy by the aggregated mask $M_t$ gives the masked conditional entropy image. The \gls{mce} loss maximizes the percentage of the masked conditional spatial entropy.}
    \label{fig:conditional}
    \vspace{-0.6cm}
\end{figure}
\noindent The \textbf{\gls{mae} loss} \label{sec:me}
 encourages maximum information coverage by the keypoints in a single frame. We use the heatmap $h_t^{(i)}$ of each keypoint $k_t^{(i)}$ to retrieve the local image information at time $t$.
We filter out inactive keypoints by
multiplying the heatmap with the activation status $s_t^{(i)}$. Aggregating all heatmaps gives the aggregated mask $M_t=\min(\sum_i^{K} {h_t^{(i)} \odot s_i^{(t)}},1)$ (cf.~\cref{fig:infomax}). 
%  Finally, by summing the pixel-wise product of $M_t$ with the entropy image $E(I_t)$ over all pixel locations $(x,y)$, we obtain the information transferred by all keypoints. 
With this masking approach, we can consider keypoints as channels of local information, and thus, we arrive at the following proposition that bounds the spatial information loss by the keypoints' masking of the original image. The bound follows Fano's inequality \citep{sabuncu2006entropy,scarlett2019introductory}, and proves that maximizing the keypoints' masked spatial entropy indeed lowers the probability of error of the information loss by this keypoint representation.
\begin{proposition}\label{propos:max_entropy}
 Let $I_t^{M}$ be the masked image at time $t$, obtained by the operation $I_t^{M}=I_t \odot M_{t}$, where $\odot$ denotes the Hadamard (\ie, element-wise) product. Let $\mathcal{B}$ be the ``vocabulary'' of pixel intensities, and we assume that every pixel in location $(x,y)$ is uniform on $\mathcal{B}$. The \textit{average} error probability $\bar P_{\varepsilon}$ over all pixels $N=\mathrm{H} \times \mathrm{W}$ of the spatial information approximated by $I_t^{M}$ \wrt\ to the original image $I_t$ can be lower bounded by
 \begin{equation}\label{eq:bound_me}
     \bar P_{\varepsilon} \geq 1 - \frac{\sum_{x,y}(\mathcal{H}(I_t^{M}(x,y)))}{N \log{|\mathcal{B}|}} - \frac{\log2}{\log{\mathcal{|B|}}}\ .
 \end{equation}
\end{proposition}
Proof in~\cref{app:proof_max_entropy}. We can assume that the upper bound for the error probability remains 1, because of the activation $s$ of the keypoints, there is a probability that the masked image is ``empty'', \ie, all keypoints inactive. From \cref{eq:bound_me} we can see that the \gls{mae} maximization lowers the probability of error. This motivates the practical implementation of the \gls{mae} loss $ \mathcal{L}_{\text{\textit{ME}}}(I_t)$ that optimizes the percentage of the masked entropy over all pixel locations $(x,y)$, $\sum_{x,y} \mathcal{H}(I_t) \odot M_t$ \wrt\ the total image entropy $\sum_{x,y} \mathcal{H}(I_t)$
\begin{align}
    \mathcal{L}_{\text{\textit{ME}}}(I_t) &= 1 - \frac{\sum_{x,y} \mathcal{H}(I_t) \odot M_t}{\sum_{x,y} \mathcal{H}(I_t)} \\ \nonumber
    &= 1 - \frac{\sum_{x,y} \mathcal{H}(I_t) \odot \min(\sum_{i=1}^{K} {h_t^{(i)} \odot s_t^{(i)}},1)}{\sum_{x,y} \mathcal{H}(I_t)} \ .
\end{align}

\noindent The \textbf{\gls{mce} loss} encourages the keypoints to pay special attention to dynamic entities when the available number of keypoints is insufficient for covering the information in a sequence of frames.
The conditional entropy of an image $I_{t}$ at time $t$ given a reference image $I_{t-1}$ at time $t-1$ measures the information change of pixels, indicating moved objects. Optimizing the conditional entropy $\mathcal{H}(I_{t}|I_{t-1})$
in a sequence of images encourages the keypoint detector to attend to moving objects (cf.~\cref{fig:conditional}). 
The spatial conditional entropy can be computed by subtracting the reference image entropy from the joint entropy of two images  $\mathcal{H}(I_{t}|I_{t-1})=\mathcal{H}(I_{t},I_{t-1})-\mathcal{H}(I_{t-1})$, where $\mathcal{H}(I_{t},I_{t-1})\approx\max(\mathcal{H}(I_{t}),\mathcal{H}(I_{t-1}))$ following

\begin{lemma}\label{lemma:joint_entropy}
The joint spatial entropy of two images $I_1$ and $I_2$ can be approximated by $\mathcal{H}(I_1(x,y),I_2(x,y)) \approx max(\mathcal{H}(I_1(x,y)),\mathcal{H}(I_2(x,y))), \forall (x,y) $, since the per pixel maximum of the marginal entropies is a lower bound of the joint entropy.
\end{lemma}
\vspace{-0.1cm}
Proof in~\cref{app:proof_joint_entropy}. Accordingly, we can bound the information loss by the keypoints in a sequence of frames.

\begin{corollary}\label{corol:cond_entropy}
Following \cref{propos:max_entropy}, we can bound the average probability of error $\bar P_{\varepsilon}^{\text{cond}}$ of the conditioned masked images between timestep $t-1$ and $t$ as
\begin{equation}\label{eq:cond_entropy}
\bar P_{\varepsilon}^{\text{cond}} \geq 1 - \frac{\sum_{x,y}\mathcal{H}(I_t^{M}(x,y)|I_{t-1}^{M}(x,y) )}{N \log{|\mathcal{B}|}} - \frac{\log2}{\log{\mathcal{|B|}}}\ .
\end{equation}
\end{corollary}

Following \cref{eq:cond_entropy}, we observe that the \gls{mce} maximization lowers the probability of error of the conditional spatial information loss between frames, leading to the practical implementation of the \gls{mce} loss, similarly to the \gls{mae} loss. The \gls{mce} loss $\mathcal{L}_{\text{\textit{MCE}}}(I_t,I_{t-1})$ maximizes the percentage of total masked conditional entropy $\sum_{x,y} \mathcal{H}(I_t|I_{t-1}) \odot M_t$ to the total conditional entropy $\sum_{x,y} \mathcal{H}(I_t|I_{t-1})$
\begin{equation}
    \mathcal{L}_{\text{\textit{MCE}}}(I_t,I_{t-1}) = 1 - \frac{\sum_{x,y} \mathcal{H}(I_t|I_{t-1}) \odot M_t}{\sum_{x,y} \mathcal{H}(I_t|I_{t-1})} \ .
\end{equation}
% \begin{lemma}\label{lemma:combined_me_mce}
% Maximizing both masked image entropy and masked conditional entropy minimizes the overall probability of error of transmitting the information of image $I_t$, as
% \begin{equation}
%     \bar P_{\varepsilon, \varepsilon^{\text{cond}}} \geq 1 -\frac{\sum_{x,y}E(I_t^{M}(x,y))+E(I_t^{M}(x,y)|I_{t-1}^{M}(x,y))}{2N \log{|\mathcal{B}|}} - \frac{\log2}{\log{\mathcal{B}}}.
% \end{equation}
% \end{lemma}
% where $CE(I_t|I_{t-1}) = max(E(I_t),E(I_{t-1})) - E(I_{t-1})$ is the pixel-wise conditional entropy.% computation using the local entropy layer.

\subsubsection{Maximizing Keypoint Information Transportation} \label{sec:it}
\begin{figure}[t!]
\centering
    \includegraphics[width=\linewidth]{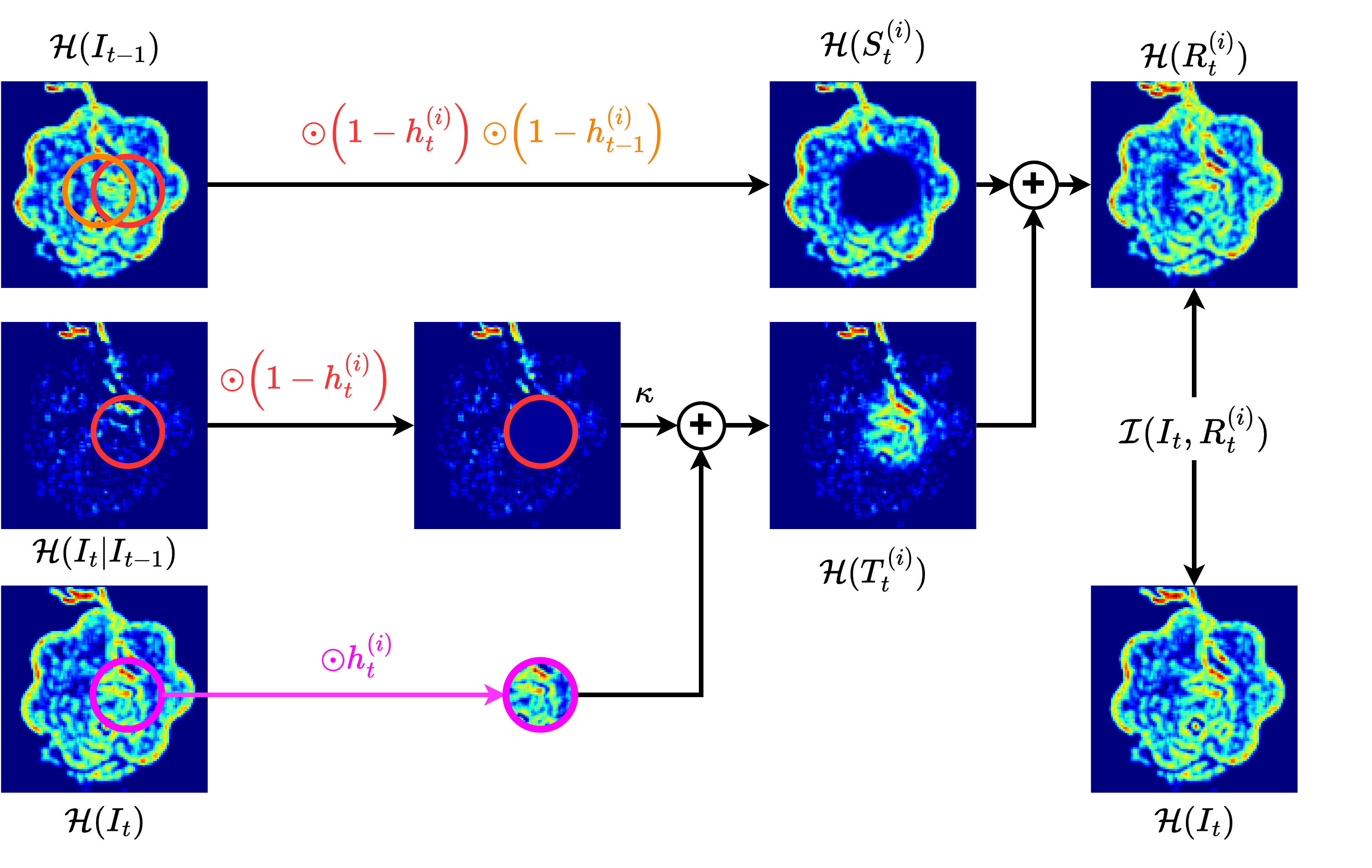}
    \vspace{-0.6cm}
    \caption{\gls{it} for keypoint $k^{(i)}$.
% The source is the spatial entropy of the previous image $\mathcal{H}(I_{t-1})$ and the target is the spatial entropy of the current frame $\mathcal{H}(I_{t})$.
 Removing the heatmap masks of the $i^{\text{th}}$ keypoint at times $t-1$ (\textcolor{orange}{orange circle}) and $t$ (\textcolor{red}{red cricle}) from the spatial entropy $\mathcal{H}(I_{t-1})$ of the image $I_{t-1}$ gives the source entropy $\mathcal{H}(S_t^{(i)})$. 
 Implanting the local entropy of the keypoint at time $t$ from the current frame $\mathcal{H}(I_{t})$ (\textcolor{magenta}{magenta circle}) into the conditional entropy $\mathcal{H}(I_{t}|I_{t-1})$ (weighted by $\kappa$) 
 after removing the heatmap mask of the keypoint at time $t$ (\textcolor{red}{red cricles}) 
 gives the target spatial entropy $\mathcal{H}(T_t^{(i)})$. The reconstructed information after transportation $\mathcal{H}(R_t^{(i)})$ is the sum of target and source entropy. The objective of \gls{it} is to maximize the mutual information between the reconstructed entropy and the entropy of the current frame $\mathcal{I}(I_{t}, R_t^{(i)})$.}
    \label{fig:infotrans}
    \vspace{-0.7cm}
\end{figure}
% The \textbf{\gls{it}} ensures the spatio-temporal consistency of the keypoint representation by solving a keypoint transportation problem. 
Keypoints should transmit information about the same entity over time. Temporal consistency means aligning each keypoint to the same information pattern across its occurrences. Thus, we propose the operation of \glsfirst{it} based on \gls{ise}, contrarily to methods that rely on image reconstruction performing feature transportation~\cite{transporter}. 

In a temporal sequence of frames, we can perform keypoint \gls{it} by \textit{reconstructing the image spatial entropy} of the current frame $\mathcal{H}(I_t)$ using the image entropy of the previous frame $\mathcal{H}(I_{t-1})$ (cf.~\cref{fig:infotrans}). 
% The conditional spatial entropy of the two frames $\mathcal{H}(I_t|I_{t-1})$, represents the amount of pixel-wise information needed to quantify the information of $\mathcal{H}(I_t)$ given $\mathcal{H}(I_{t-1})$.
Let's consider the $i^{\text{th}}$ keypoint at time step $t$ (coordinates are omitted for avoiding verbosity). Its associated heatmap $h_t^{(i)}$ is a mask on the entropy image that allows localizing %representing 
the spatial information conveyed by the $i^{th}$ keypoint. We can construct a \textit{source entropy image} $\mathcal{H}(S^{(i)}_t)$ by subtracting the local entropy of the $i^{th}$ keypoint in frames $t-1$ and $t$ from the entropy image $\mathcal{H}(I_{t-1})$, \ie, $\mathcal{H}(S^{(i)}_t)=\mathcal{H}(I_{t-1})\odot(1-h_{t-1}^{(i)})\odot(1-h_t^{(i)})$. 
The conditional spatial entropy of the two frames $\mathcal{H}(I_t|I_{t-1})$ represents the amount of pixel-wise information needed to quantify the information of $\mathcal{H}(I_t)$ given $\mathcal{H}(I_{t-1})$.
Implanting the keypoint's spatial entropy covered by $h_t^{(i)}$ onto the conditional image entropy $\mathcal{H}(I_t|I_{t-1})$, that contains all conditional information except for the information transmitted by the $i^{\text{th}}$ keypoint, forms the target image entropy $\mathcal{H}(T^{(i)}_t)=\mathcal{H}(I_{t}) \odot (h_t^{(i)})+\kappa \mathcal{H}(I_{t}|I_{t-1}) \odot (1-h_t^{(i)}).$\footnote{The factor $\kappa \leq 1$ encourages the network to concentrate more on transportation than reconstruction.} The reconstruction of the image entropy $\mathcal{H}(I_t)$ results from the pixel-wise sum of the source and target image entropies $\mathcal{H}(R^{(i)}_t)=\mathcal{H}(S^{(i)}_t)+\mathcal{H}(T^{(i)}_t)$. The transportation loss is computed independently per keypoint, and enforces each keypoint to consistently represent the same information pattern spatio-temporally. The reconstruction process of our \gls{it} leads us to the following proposition, showing that maximizing the \gls{mi} between the per keypoint reconstructed information and the original image entropy lowers the probability of error due to information loss. 
\begin{proposition}\label{propos:it}
 \setlength{\abovedisplayskip}{3pt}
\setlength{\belowdisplayskip}{3pt}
Following Fano's inequality \citep{sabuncu2006entropy,scarlett2019introductory}, we prove that the average error probability of the transportation of the $i^{\text{th}}$ keypoint $P_{\varepsilon}^{\text{IT} (i)}$, assuming each keypoint transportation independently, is lower bounded by
\begin{equation}\label{eq:bound_mi}
 \setlength{\abovedisplayskip}{3pt}
\setlength{\belowdisplayskip}{3pt}
    \bar P_{\varepsilon}^{\text{IT} (i)} \geq 1 - \frac{\sum_{x,y}\mathcal{I}(I_t(x,y),R^{(i)}_{t}(x,y))}{N \log{|\mathcal{B}|}} - \frac{\log2}{\log{\mathcal{B}}} \ .
\end{equation}
% Assuming all transportations independent, we can bound the joint average error probability, by summing the per keypoint error probability 
% \begin{equation}\label{eq:bound_mi_tot}
%      \bar P_{\varepsilon}^{\text{IT (joint)}} \geq K - \frac{\sum_{i=1}^{K}\sum_{x,y}\mathcal{I}(I_t(x,y),R^{(i)}_{t}(x,y))}{ N \log{|\mathcal{B}|}} - \frac{K\log2}{\log{\mathcal{B}}}.
% \end{equation}
\end{proposition}
Proof in~\cref{app:proof_it}.
% As the heatmaps represent the information coverage of keypoints, the reconstruction process uses them from both time frames ($h_i^{(t)}$ and $h_i^{(t-1)}$), and removes them from the previous frame information to form the source information $E(S_t)=E(I_{t-1})\odot(1-h_i^{(t-1)})\odot(1-h_i^{(t)})$.
From \cref{eq:bound_mi}, we deduce that for optimizing the $i^{\text{th}}$ keypoint's \gls{it}, we should maximize the \gls{mi} $\mathcal{I}(I_t, R^{(i)}_t) $. This motivates our practical implementation of the \gls{it} loss for all keypoints, and we construct the \gls{it} loss through the difference $\mathcal{H}(I_{t}) - \mathcal{I}(I_{t},R_t^{(i)})$ normalized by the area of the heatmap $A_{h}$ (equal for all keypoints). Minimizing $\mathcal{H}(I_{t}) - \mathcal{I}(I_{t}, R_t^{(i)})$ maximizes \gls{mi}, as dictated by \cref{propos:it}. We found that normalizing with $A_{h}$ helps having a better loss scale. We also regularize the excessive keypoint movement by minimizing the norm of the distance traveled by each keypoint $d_t^{(i)}=||(x_i,y_i)_t-(x_i,y_i)_{t-1}||_2^2$ (scaled by a weight $m_d$). The practical implementation of the \gls*{it} loss for all keypoints becomes
\begin{equation}\label{eq:infotrans}
        \mathcal{L}_{\text{IT}}(I_t,I_{t-1})=\sum_{i=1}^K \frac{\sum_{x,y}\mathcal{H}(I_{t}) - \mathcal{I}(I_{t},R_t^{(i)})}{A_{h}} +m_d \cdot d_t^{(i)}
        % + ||(x_i,y_i)_t-(x_i,y_i)_{t-1}||_2^2
        \ .
    \end{equation}
% where the mutual information computed as $\mathcal{I}(R_t,I_t) = E(R_t) + E(I_{t}) - E(R_t,I_{t})$, with the joint entropy approximated by $E(R_t,I_{t}) = max(E(R_t),E(I_{t}))$. 
% \subsubsection{The \gls*{mint} loss \& Auxiliary losses}
\subsubsection{The \gls{mint} loss \& Auxiliary losses}
\begin{figure*}[t!]
    \centering
    \includegraphics[width=0.9\linewidth]{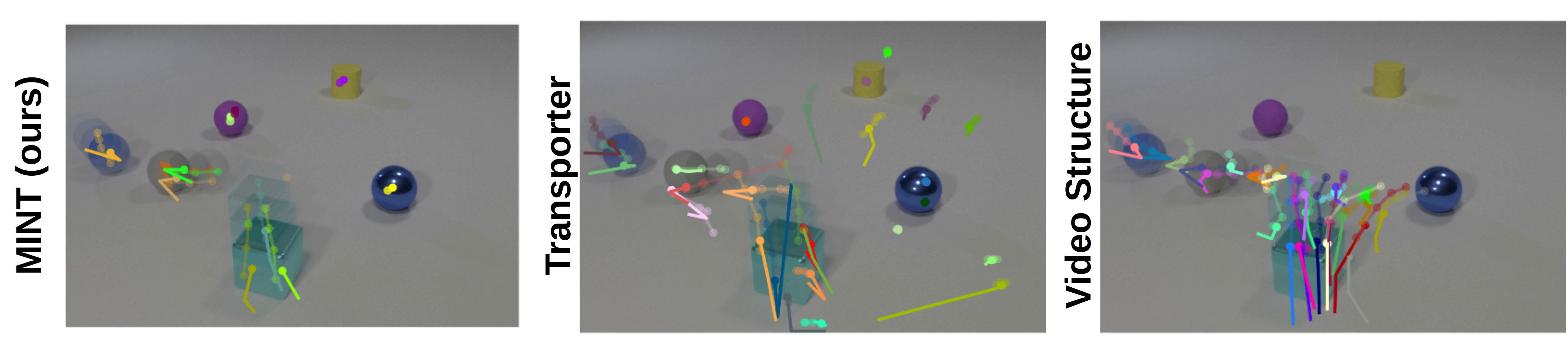}
    \vspace{-0.5cm}
    \caption{Qualitative results on CLEVRER for Task I (object detection and tracking) and Task II (learning dynamics). Our method is able to assign keypoints to all objects, independently of whether they move or not, and follows their trajectory. The number of keypoints is dynamically adjusted to the number of objects.
    Future states, the predicted keypoint and trajectories, are transparent. 
    }
    \label{fig:objects}
    \vspace{-0.2cm}
      \captionof{table}{Quantitative evaluation of keypoint detection and tracking on CLEVRER \citep{clevrer}.}
    %   The metrics are described in \cref{sec:obj}.}
      \label{table:comparison}
      \footnotesize
      \centering
      \adjustbox{max width=0.75\linewidth}{%
      \begin{tabular}{ccccc}
        \toprule
        Method & \textbf{\acrshort*{dop} $\Uparrow$} & \textbf{\acrshort*{top} $\Uparrow$} &\textbf{\acrshort*{uak} $\Downarrow$}& \textbf{\acrshort*{rak} $\Downarrow$}\\
        \midrule
        MINT w/o Reg. (ours) & \textbf{0.918 $\pm$ 0.073} & \textbf{0.897 $\pm$ 0.078} & 6.793 $\pm$ 1.956 & 2.478 $\pm$ 0.865\\
        % MINT static (ours) & 0.849 $\pm$ 0.115 & 0.826 $\pm$ 0.119 & 0.958 $\pm$ 0.615 & 
        % 1.142 $\pm$ 0.446 \\
        % 3.137 $\pm$ 1.034 \\
        % MINT (ours) &  \textbf{0.847 $\pm$ 0.125} & \textbf{0.827 $\pm$ 0.129} & \textbf{1.864 $\pm$ 0.876} & \textbf{0.999 $\pm$ 0.385}\\
        MINT (ours) &  \textit{0.855 $\pm$ 0.118} & \textit{0.838 $\pm$ 0.121} & \textbf{0.889 $\pm$ 0.639} & 
        \textbf{1.123 $\pm$ 0.448}\\
        % \textbf{3.022 $\pm$ 1.156}\\
        Transporter & 0.787 $\pm$ 0.113 & 0.745 $\pm$ 0.119 & 18.417 $\pm$ 1.639 & 
        1.157 $\pm$ 0.323\\
        % 5.191 $\pm$ 2.122\\
        Transporter-modified & 0.832 $\pm$ 0.107 & 0.794 $\pm$ 0.114 & 16.267 $\pm$ 2.349 & 
        1.764 $\pm$ 0.671\\
        % 5.191 $\pm$ 2.122\\
        Video Structure  & 0.567 $\pm$ 0.256 & 0.543 $\pm$ 0.253 & 18.104 $\pm$ 3.538 & 
        1.922 $\pm$ 0.652 \\
        % 4.531 $\pm$ 2.833 \\
        \bottomrule
      \end{tabular}}
            \vspace{-0.23cm}
      \captionof{table}{Prediction success rate on CLEVRER \citep{clevrer}.}
      \label{table:dynamics}
      \footnotesize
      \centering
      \adjustbox{max width=0.75\linewidth}{%
      \begin{tabular}{cccc}
        \toprule
        Method & 1-step prediction & 2-steps prediction & 3-steps prediction \\
        \midrule
        % MINT w/o Reg. (ours) & \textit{0.916 $\pm$ 0.063} & \textit{0.907 $\pm$0.068} & \textit{0.889 $\pm$0.076} \\
        MINT (ours) & \textbf{0.844 $\pm$ 0.116} & \textbf{0.827 $\pm$0.126} & \textbf{0.811$\pm$0.132} \\
        Transporter & 0.746 $\pm$ 0.116 & 0.716 $\pm$ 0.120 & 0.692 $\pm$ 0.122\\
        Transporter-modified & 0.814 $\pm$ 0.099 & 0.791 $\pm$ 0.106 & 0.769 $\pm$ 0.110\\
        Video Structure  & 0.734 $\pm$ 0.124 & 0.719 $\pm$ 0.125 & 0.699 $\pm$ 0.127 \\
        \bottomrule
      \end{tabular}}
    \vspace{-0.5cm}
    \end{figure*}
    
\noindent The~\textbf{overlapping loss} provides an auxiliary supervisory signal that spreads the keypoints over the image, encouraging them to cover distinctive regions. The sum of the Gaussians $G_t^{(i)}$ (cf. \cref{sec:heatmap}) around the keypoints $k_t^{(i)}$ helps to estimate their overlap. The overlapping loss,  % $\mathcal{L}_o= min(max(\sum^{K}_i G_t^{(i)})-\beta,0)/K$, 
\begin{equation}
        \mathcal{L}_o= \frac{1}{K}\min(\max(\sum^{K}_i G_t^{(i)})-\beta,0)\ ,
\end{equation}
minimizes the maximum of the aggregated Gaussians normalized by the number of keypoints $K$ with a lower bound $\beta$ to allow some occlusions and avoid over-penalization.

\noindent The~\textbf{active status loss} encourages the model to deactivate unnecessary keypoints, \ie, setting the status $s_t$ to 0, by minimizing the normalized sum of active keypoints while maximizing \gls{mae}. The interplay of the losses allows the method to eventually reach a trade-off between the number of active keypoints and covered spatial entropy. \\
The active status loss optimizes 
% $\mathcal{L}_s= (\sum_i^{K} s_t^{(i)} )/ K$.
\begin{equation}\label{eq:status_loss}
        \mathcal{L}_s= \frac{1}{K}\sum_i^{K} s_t^{(i)}
        \ .
    \end{equation}
The overall \textbf{\gls{mint} loss} $\mathcal{L}_{\text{\textit{MINT}}}$ is a weighted combination of all losses (with a dedicated weight $\lambda$ per loss), with the weight of the status loss reversed to schedule it according to the percentage of \gls{mae},
% $\mathcal{L}_{\text{\textit{MINT}}}= \lambda_{\text{\textit{ME}}} \mathcal{L}_{\text{\textit{ME}}} +%         \lambda_{\text{\textit{MCE}}} \mathcal{L}_{\text{\textit{MCE}}}+\lambda_{\text{\textit{IT}}} \mathcal{L}_{\text{\textit{IT}}}+ \lambda_{o} \mathcal{L}_{o}  + (1-\mathcal{L}_{\textit{ME}}) \lambda_{s} \mathcal{L}_{s}$. 
    \begin{align}\label{eq:mint}
    \begin{split}
        \mathcal{L}_{\text{\textit{MINT}}}= & \lambda_{\text{\textit{ME}}} \mathcal{L}_{\text{\textit{ME}}} +
        \lambda_{\text{\textit{MCE}}} \mathcal{L}_{\text{\textit{MCE}}}+\lambda_{\text{\textit{IT}}} \mathcal{L}_{\text{\textit{IT}}}\\&+\lambda_{o} \mathcal{L}_{o}  + (1-\mathcal{L}_{\textit{ME}}) \lambda_{s} \mathcal{L}_{s}  \ .
    \end{split}
    \end{align}
Further information about the hyperparameters are available in \cref{sec:hyperparameters}.

\section{Experiments}\label{sec:exper}
We evaluate \gls{mint} on four datasets ranging from videos of synthetic objects -- CLEVRER \citep{clevrer} and MAGICAL \citep{magical} -- to realistic human video demonstrations -- MIME \citep{mime} and SIMITATE \citep{simitate}. Our experiments show the efficacy of our method as a representation for different tasks, and we provide quantitative results \wrt\ evaluation
% our newly proposed 
metrics (for object detection and tracking on CLEVRER) for several downstream tasks (learning dynamics on CLEVRER, imitation learning on MAGICAL), and  qualitative results on the challenging datasets of MIME and SIMITATE.
% , demonstrating the superior suitability of our method for real-world deployment. 

We compare against baselines for unsupervised end-to-end keypoint representation learning from \emph{videos}. To the best of our knowledge, 
%As we focus on single end-to-end deep models like \gls{mint}, 
the only baselines in this context (cf. \cref{sec:related_main}) are \textbf{Transporter}~\citep{transporter} and \textbf{Video Structure}~\citep{videostructre}. Additionally, we include \textbf{Transporter-modified}, a modified version with a smaller receptive field that we designed for comparison. Further, we compare to \gls{mint} without the regularization terms (\textbf{MINT w/o Reg.}), and an end-to-end CNN-based feature extraction. We report statistics for all quantitative results over 5 seeds. % to show the consistency of~\gls{mint}.
An extensive ablation study of \gls{mint} is provided in \cref{sec:ablation} and baselines are discussed in \cref{sec:baselines}. 
%\myparagraph{DOWNSTREAM TASK I:\textit{ OBJECT DETECTION AND TRACKING}.}%\subsection{DOWNSTREAM TASK I:\textit{ OBJECT DETECTION AND TRACKING}}

\myparagraph{Downstream task I:\textit{ Object detection and tracking}.}
% ~Successful keypoint discovery corresponds to the model's ability to predict keypoints that describe the scene's structure and dynamics.
~Capturing scene structure requires detecting all objects in an image, while object tracking is essential for representing the scene's dynamics.
\gls{mint} can successfully train a spatio-temporally consistent keypoint representation on videos, leading to its natural application for object (static/dynamic, appearing/disappearing) detection and tracking. 
\\
We use CLEVRER~\citep{clevrer}, a dataset for visual reasoning with complete object annotations, containing videos with static and dynamic objects, with good variability in
% structure and dynamics of the 
scenes, as a testbed.
To quantitatively assess the performance of \gls{mint}, % for object detection and tracking, 
we developed evaluation metrics for CLEVRER. We propose the \textbf{\gls{dop}} and the \textbf{\gls{top}} as two metrics, with higher values corresponding to better keypoint detection and tracking. A keypoint detects an object if it lies on its mask, and tracks it, if it detects the same object in two consecutive frame. Assigning keypoints to areas already represented by other keypoints or empty spaces signals bad keypoint detection. To evaluate these cases, we define two additional metrics for the \textbf{\gls{rak}} and \textbf{\gls{uak}}, with lower values corresponding to better detection. The metrics are described in detail in \cref{sec:metrics}.
\\
We train all keypoint detectors on a subset of 20 videos from CLEVRER and test them on 100. The train-test split emulates a low-data regime and tests the methods' generalization abilities. 
% Not only our method outperforms the baselines (\cref{table:comparison}), but it is also faster and more efficient to train.
% thanks to dropping the need for reconstruction.
As seen in~\cref{table:comparison},
MINT w/o Reg. detects more objects (\gls{dop}) and tracks them better (\gls{top}), showing the benefit of our information-theoretic losses. The proposed 
\gls{mint} model exhibits the best trade-off between superior performance against the baselines on all metrics, and better handling of keypoint assignment (\gls{uak} and \gls{rak}) than MINT w/o Reg. This is due to the computation of the supervisory entropy signal being overestimated, but the regularizers balance this effect. See  visual comparisons in \cref{fig:objects} or in the video results,\footnote{Videos on the website \url{https://sites.google.com/view/mint-kp}.} and more discussion about the ablations in \cref{sec:ablation}.
% economizes the number of keypoints (\gls{uak}), and reduces redundant keypoint assignment (\gls{rak}) while outperforming the baselines.
% Despite being trained on single images, MINT static performs very well on relevant metrics. 
% As seen in~\cref{table:comparison}, \gls{mint} detect more objects (\gls{dop}) and tracks them better (\gls{top}). Meanwhile, the overall \gls{mint} economizes the number of keypoints (\gls{uak}), while reducing redundant keypoint assignment (\gls{rak}). 

%Also the visual results in \cref{fig:objects} demonstrate the efficacy of our method in learning superior keypoints. % apart from our quantitative analysis. 
% \gls{mint} detects and tracks the objects in the scene, and its keypoints are more interpretable than the baselines, allowing better scene understanding.

%%%%%%%%%%%%%%%%%%%%%%%%%%%%%%%%%%%%%%%%%%%%%%%%%%%%%%%%%%%%%%%%%%%%%%%%%%
%\myparagraph{DOWNSTREAM TASK II: \textit{LEARNING DYNAMICS}.}
\myparagraph{Downstream task II: \textit{Learning dynamics}.}
\begin{figure}[t!]
\centering
\includegraphics[width=\linewidth]{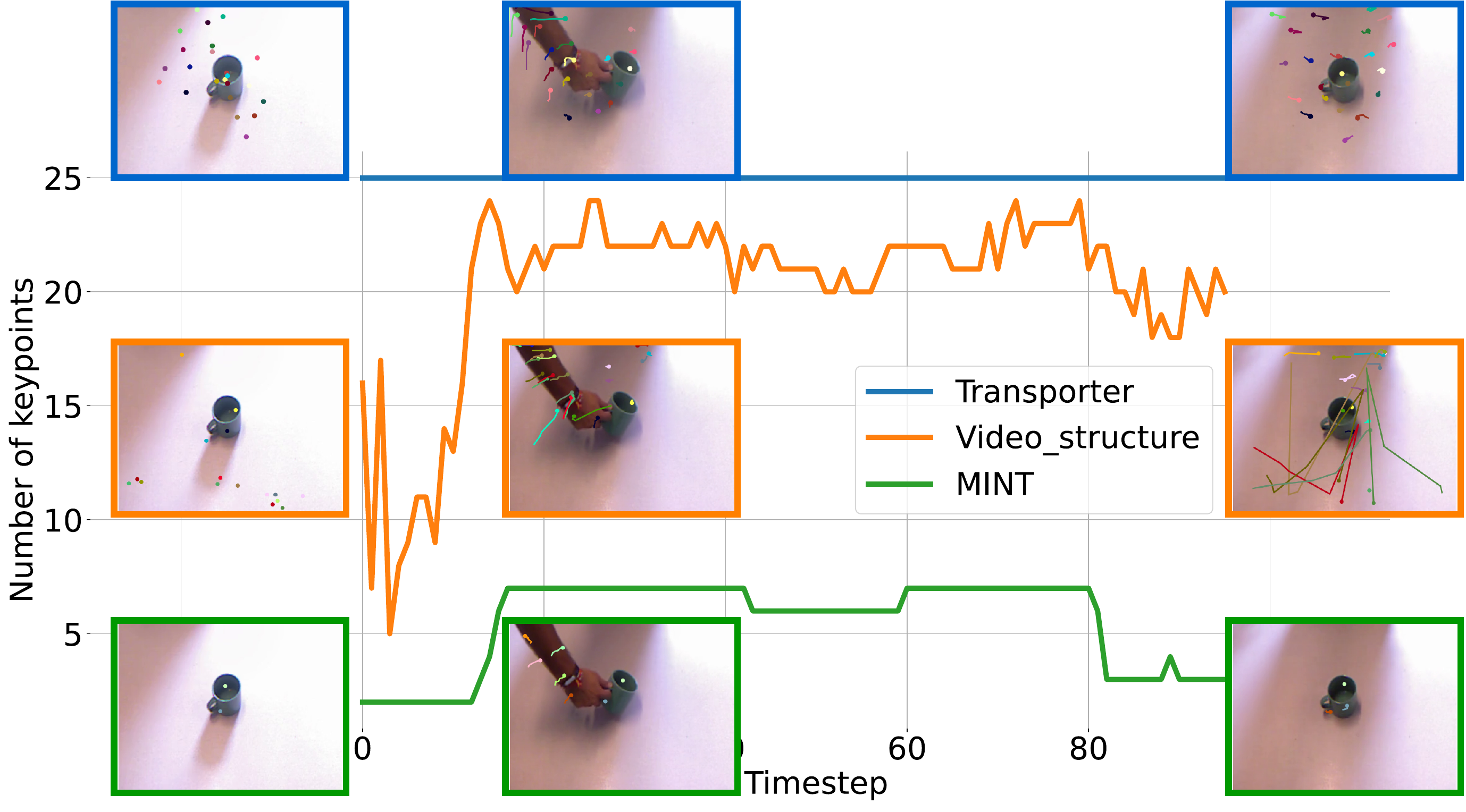}
    \vspace{-0.38cm}
    \caption{Come-and-go scenario in MIME. The hand enters after the start of the video and departs before the end. We plot the number of active keypoints \wrt\ time. Transporter ~\citep{transporter} has a fixed number of keypoints. Video structure ~\citep{videostructre} increases the number of active keypoints when the hand appears, but struggles when it disappears. \gls{mint} uses a suitable number of keypoints.}
  \label{fig:comeandgo}
  \vspace{-0.6cm}
\end{figure}
~Proper object detection allows us to learn the underlying dynamics that evolve in a scene.
% , enabling the prediction of the next stage of the scene. 
We test the representation power of the discovered keypoints by training a prediction model
(\ie, a model predicting the next state of the objects)
using the pre-trained keypoint detectors from Task I (using the best seed for each method). The prediction model treats the keypoints as graph nodes in an \gls{in} \citep{battaglia2016interaction} to model the relational dynamics (cf. \cref{a_interact}). We train the prediction model to forecast the future positions of the keypoints given a history of four-time steps. We compare the prediction against the ground truth position of the object in the predicted frame using CLEVRER~\citep{clevrer}. We report in \cref{table:dynamics} the ratio of successfully predicted objects (\ie, a predicted keypoint lying on the same object in the next frame) to the ground truth number of objects in the next time step. The comparison demonstrates that keypoints detected by our method represent the scene better than the baselines and help to predict the next state. \cref{fig:objects} shows the prediction performance using different keypoint detectors. 
% The comparison demonstrates that keypoints detected by our method convey fidelity information about the objects' positions, velocities, and acceleration better than the baselines.

%%%%%%%%%%%%%%%%%%%%%%%%%%%%%%%%%%%%%%%%%%%%%%%%%%%%%%%%%%%%%%%%%%%%%%%%%%
%\myparagraph{DOWNSTREAM TASK III: \textit{OBJECT DISCOVERY IN REALISTIC SCENES}.} 
\begin{figure}[t!]
        \centering
        \includegraphics[trim={0cm 0 0 0.4cm},clip,width=\linewidth]{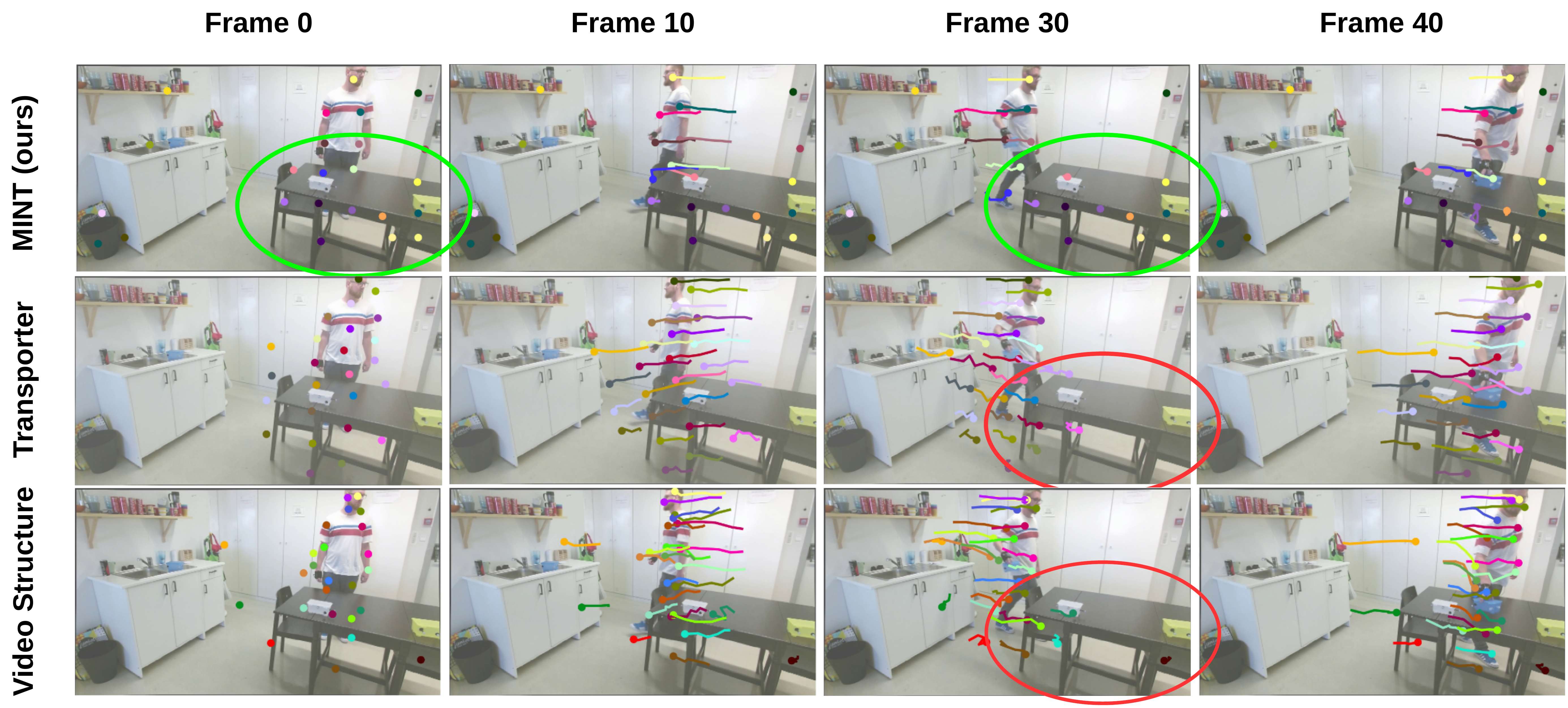}
        % \setlength{\belowcaptionskip}{-25pt}
        % \vspace{-0.39cm}
        \caption{Crowded scene from SIMITATE with a human moving in a room. All methods can track the human successfully, but only \gls{mint} can keep keypoints on the static objects consistently (green ellipses), while the baselines lose track of them (red ellipses).}
        \label{fig:simitate}
        \vspace{-0.5cm}
    \end{figure}
    
\myparagraph{Downstream task III: \textit{Object discovery in realistic scenes}.} Our method addresses challenging aspects beyond synthetic datasets. We evaluate the keypoint detectors on two additional datasets: (1) MIME \citep{mime}: a collection of close-up videos of human hands manipulating objects, and (2) SIMITATE \citep{simitate}: a video dataset of humans performing manipulation tasks in wide-view cluttered scenes. 
% For training, we use a subset of 80 videos from various tasks, and for testing 100, and 
Since no annotations are provided in these datasets, we perform only qualitative analysis.

In MIME, the human hand enters and leaves the scene abruptly, allowing to evaluate \gls{mint} in come-and-go scenarios as shown in \cref{fig:comeandgo}. \gls{mint} only activates the necessary number of keypoints, while Transporter uses a static number, and Video structure fails to deactivate the excessive keypoints when the hand disappears. \cref{fig:comeandgo} shows the number of active keypoints over time, revealing our method's superior performance for the number of keypoints and the qualitative representation of objects in the scene.
% thanks to the active status loss.
\begin{figure*}[t!]
      \captionof{table}{Average score for imitation learning on MAGICAL \citep{magical}. Higher values are better.}
      \label{table:imitation}
      \small
      \centering
      \adjustbox{max width=0.75\linewidth}{%
      \begin{tabular}{c cc cc cc}
        \toprule
        Method & \multicolumn{2}{c}{MoveToRegion} & \multicolumn{2}{c}{MoveToCorner} & \multicolumn{2}{c}{MakeLine} \\
        & Demo & TestJitter & Demo & TestJitter & Demo & TestJitter\\
        \midrule
        MINT (ours) & \textbf{1.00 $\pm$ 0.00} & \textbf{0.86 $\pm$ 0.31} & \textbf{1.00 $\pm$ 0.00} & \textbf{0.80 $\pm$ 0.34} & \textbf{0.2 $\pm$ 0.22} & 0.06 $\pm$ 0.14\\
        % Video Structure  & 1.00 $\pm$ 0.00 & 0.96 $\pm$ 0.18 & 1.00 $\pm$ 0.00 & 0.76 $\pm$ 0.37 & 0.10 $\pm$ 0.18 & 0.01 $\pm$ 0.07 \\
        % Transporter & 1.00 $\pm$ 0.00 & 0.86 $\pm$ 0.32 & 1.00 $\pm$ 0.00 & 0.82 $\pm$ 0.33 & 0.00 $\pm$ 0.01 & 0.01 $\pm$ 0.06\\
        CNN & \textbf{1.00 $\pm$ 0.00} & 0.84 $\pm$ 0.32 & 0.74 $\pm$ 0.35 & 0.30 $\pm$ 0.38 & 0.00 $\pm$ 0.00 & 0.01 $\pm$ 0.06\\
        \bottomrule
      \end{tabular}}
    \vspace{-0.4cm}
    \end{figure*}
\begin{figure}[t!]
\centering
\includegraphics[width=0.9\linewidth]{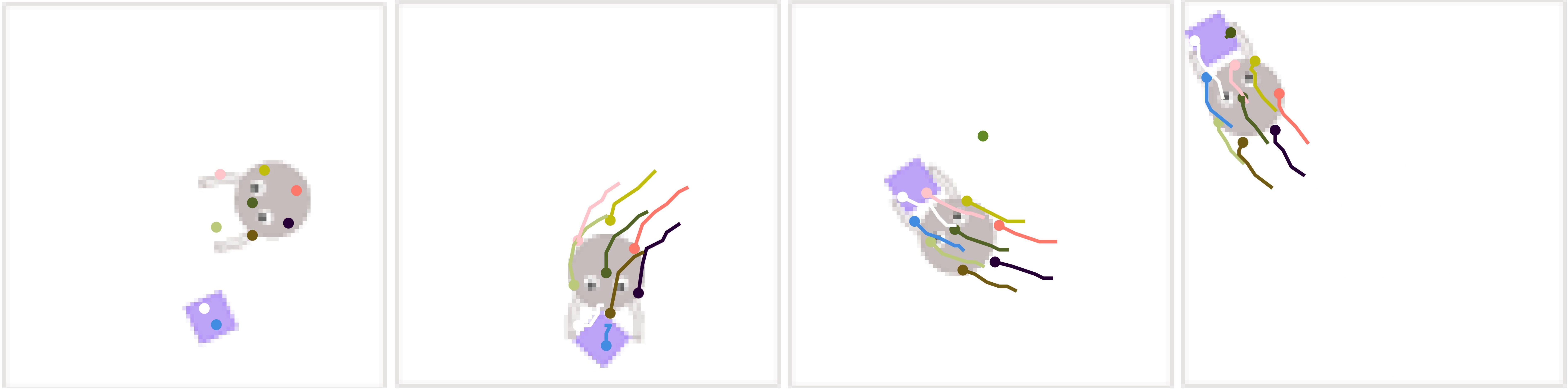}
    \vspace{-0.38cm}
    \caption{Keypoint-based imitation learning in MAGICAL. The figure showcases the MoveToCorner environment, where our agent's objective is to move the purple object to the top-left corner. Our approach, \gls{mint}, enables the agent to observe keypoints that describe the environment and predict the next action accurately. We demonstrate the effectiveness of our method by training the agent to imitate expert trajectories. The visualization overlays \gls{mint} keypoints on sample frames from a successful rollout.}
  \label{fig:magical_main}
  \vspace{-0.4cm}
\end{figure}

% SIMITATE provides a good testbed for scenarios with crowded scenes containing static and dynamic objects, with more objects than the number of available keypoints. 
The qualitative results for SIMITATE in~\cref{fig:simitate}, on the other hand, show that only \gls{mint} can disambiguate between static and dynamic objects, tracking human movement, while maintaining the structure of the keypoints relatively constant over the static objects. The baselines rely on reconstructing the movement, failing to represent the scene's structure. The qualitative results reveal the need for the conditional entropy loss (forcing attention on moving objects when the number of available keypoints is restricted) and the information transportation loss (ensuring the spatio-temporal consistency).
% -- see  also \cref{fig:simitateapp} in Appendix.
We further include ablation study results on both realistic datasets in \cref{sec:ablation}.
% , where we show and discuss the effect of regularization (comparing to MINT w/o Reg.) and temporal information (comparing to MINT w/o Temp.) qualitatively.

%%%%%%%%%%%%%%%%%%%%%%%%%%%%%%%%%%%%%%%%%%%%%%%%%%%%%%%%%%%%%%%%%%%%%%%%%%
%\myparagraph{DOWNSTREAM TASK IV: \textit{IMITATION LEARNING}.}
\myparagraph{Downstream task IV: \textit{Imitation learning}.}
Imitation learning from video frames is a long-standing challenge for control.
% , where every possible combination of pixel values can define a state.
Keypoints can define a low-dimensional representation that could reduce the computational burden considerably. In this experiment, we investigate the suitability of our keypoint representation for control tasks, like imitation learning in MAGICAL~\citep{magical} (cf. \cref{fig:magical_main}). We first pretrain \gls{mint} on 24 demonstration videos from different tasks. Then, we fix the keypoint detector and train an agent to mimic the demonstrated actions, using an \gls{in} \citep{battaglia2016interaction}, followed by a fully-connected layer that decodes the actions (cf. \cref{a_interact}). The agent uses as input the observed keypoints from four frames. We also found it useful to predict the next state as an auxiliary task. We compare the \gls{mint}-based agent against an agent that uses a CNN to extract features directly from pixels. The CNN agent is trained from scratch for each environment (cf.~\cref{sec:a_magical}). We consider three environments with different levels of difficulty; \textbf{MoveToRegion}: move an agent to a specific region, only the agent is involved (easy). \textbf{MoveToCorner}: move an object to the top-left corner, one object and the agent are involved (medium). \textbf{MakeLine}: place multiple objects in a line, four objects and the agent are involved (hard). We evaluate the learned policy on environment instances from demonstrations (Demo) and randomly initialized (TestJitter). The results in \cref{table:imitation} reveal that a pretrained keypoint model with \gls{mint} is suitable for control, achieving comparable or even superior performance to a task-specific CNN-based agent (cf. \cref{sec:a_magical} for more details).\\
We hypothesize that there is still room for improvement to unleash the potential of \gls{mint} sparse keypoint representation for control. One viable option is to use a more expressive network architecture (e.g., graph attention network) that may provide a better representation of the keypoint-induced graph. Another promising direction is to boost a reinforcement learning agent with the imitation learning policy. However, this is out of the scope of the current work.

% We hypothesize that the sparse \gls{mint} keypoint representation may require a different network architecture (e.g., a graph attention network) to better encode relations needed for learning this task, or that the available demonstrations are not enough to provide a policy with sufficient support. A viable solution would be to use the imitation learning policy as a boosting for a reinforcement learning agent, however, this is out of the scope of the current work.

\myparagraph{Limitations.} Our method relies on \gls{ise} after filtering high-frequency color changes. As a result, the method has difficulties in recognizing transparent objects and objects with the same color as the background. We plan to investigate the integration of implicit representation learning to counteract this issue. 
Another limitation is the interpretation of the keypoints in the three-dimensional space. The current method operates on images and does not provide 3D information. Adding depth information or extending to a multi-view setting are options for future improvements. 
Our method can use high-level features from a pretrained encoder to estimate the entropy, which may solve the current limitations. However, while we treated pixels as discrete random variables with RGB values, high-level features lie on a continuous latent space and, therefore, would require variational inference techniques. This direction would require a new treatment compared to our current analysis that relies on discrete probability theory, which goes beyond the scope of the current paper that lays the ground for such future work.
% We considered using high-level features from a pretrained encoder to estimate the entropy. However, our losses are not designed for this domain, nor did we have this in mind when deriving our method. While with RGB values, we have a treatment of pixels as discrete random variables, high-level features lie on a continuous latent space and, therefore, would require variational inference techniques, which would require a new treatment compared to our current analysis that relies on discrete probability theory. A potential solution to this issue is to discretize the latent space via vector quantization to compute the entropy through our entropy layer, which can be adapted to operate on quantized latent spaces. This exciting research direction goes beyond the scope of the current paper that lays the ground for such future work.

\section{Related Work}  \label{sec:related_main}

\noindent \textbf{Representation learning.}
The idea of extracting sparse feature representations of high-dimensional visual data is dominant in computer vision and machine learning research \citep{harris1988combined,lowe2004distinctive}, and connects to the functioning of the human visual system \citep{marr2010vision}. Such sparse representations are generally known as \gls{poi}, which are 2D locations that are stable
and repeatable under various lighting conditions and viewpoints \citep{detone2018superpoint}. Traditional geometric computer vision methods relied on the extraction of hand-crafted feature descriptors~\citep{lowe2004distinctive,rublee2011orb} for tasks like localization \citep{schmid2000evaluation,mur2015orb}. In the deep learning era, CNN architectures have proven superior to handcrafted features~\citep{yi2016lift,detone2018superpoint,sekd,zheng2017sift}. Deep approaches extract clouds of  \gls{poi} that are useful for correspondence searching in visual place recognition from different viewpoints \citep{hausler2021patch}, or pose-estimation for control~\citep{florence2019correspondence}.
Related to our method are object-centric approaches \cite{singh2021illiterate, locatello2020object, dittadi2022generalization}, which aim to learn abstract representations for objects in a scene. Our approach to keypoint discovery, alongside our metrics, are directed at learning and evaluating keypoints as an object-centric representation.

\noindent \textbf{Image information entropy.} Our work draws inspiration from classical approaches in saliency detection in images that use local information to detect salient 
entities~\citep{kadir2001saliency,bruce2005saliency,fritz2004attentive,renninger2004information,borji2012state}. 
\citet{bruce2005saliency} proposes that regions with high self-information typically correspond to salient objects, and \citet{alexe2010object} quantified objectiveness by self-information approximated via center-surround feature differences. Extracting sparse feature representations of high-dimensional visual data is also dominant in \gls{cv} \citep{harris1988combined,lowe2004distinctive, detone2018superpoint}. 
Traditional \gls{cv} methods relied on the extraction of hand-crafted feature descriptors~\citep{lowe2004distinctive,rublee2011orb,schmid2000evaluation,mur2015orb}.  
Notably, image information entropy has additional applications in various \gls{cv} problems, like image registration \citep{sabuncu2006entropy}, active vision \citep{ferraro2002entropy}, medical image analysis \citep{hrvzic2019local}, nuclear detection \citep{hamahashi2005detection}, image compression \citep{minnen2018image}, and image randomness \citep{wu2013local}. Our method proposes information-theoretic losses based on the \gls{ise}~\citep{brink1996using,razlighi2009comparison}. The use of \gls{ise} was prevalent in \gls{cv} applications for image reconstruction \citep{gull1978image} and in Markov Random Fields~\citep{li2009markov,5200466}.  

\noindent \textbf{Keypoint learning.} Keypoints represent a class of \gls{poi} that have a semantic entity, \eg, representing objects~\citep{duan2019centernet}, or human joints~\citep{cao2017realtime,mcnally2021rethinking}, but most methods rely on explicit annotations of keypoint locations. Related to our work are unsupervised methods for keypoint detection. \citet{jakab} use an autoencoder architecture with a differentiable keypoint bottleneck trained on the difference between a source and a target image, trying to restrict the information flow. \gls*{mint} also uses a differentiable keypoint representation, but it operates on the output of an hourglass architecture. Our results suggest that learning to redistribute the information after compression is beneficial for keypoint discovery~\citep{hourglass}. \citet{videostructre} use a similar architecture as~\citet{jakab} but operate on video sequences for detecting keypoints, using the intensity of the bottleneck heatmap as an indicator of the importance of a keypoint. Setting up a threshold on the intensity is challenging and domain-specific. Contrarily, we learn a binary classification of active/inactive keypoints and optimize the number of keypoints used in every frame. \citet{transporter} propose feature transportation in the keypoint bottleneck of \citet{jakab} before reconstruction. \gls*{mint} performs information transportation and waives the need for image reconstruction, which would require an additional appearance encoder and a reconstruction decoder from keypoints. \citet{gopalakrishnan2020unsupervised} devised a three-stage architecture that first learns a spatial feature embedding, then solves a local spatial prediction task related to object permanence, and finally converts error maps into keypoints. However, this method operates on single images and does not consider the temporal consistency of the extracted keypoints. \gls{mint}, on the other hand, is a parameter-efficient single model that provides spatio-temporally consistent keypoints in videos.
% While this method employs local information, it trains three architectural modules separately, unlike \citep{transporter,videostructre} and \gls*{mint} that use a single end-to-end pipeline for keypoint discovery.
To the best of our knowledge, \citep{transporter, videostructre} are the only comparable methods that set the state-of-the-art for unsupervised temporal keypoint discovery, also proven by their successful adoption for behavior recognition \citep{sun2022behavior}, causal discovery \citep{li2020causal}, and control \citep{bechtle2023multimodal}.

In~\cref{sec:extended_related}, we also discuss 
% representation learning methods and 
the use of information-theoretic measures in deep learning.

\section{Conclusion} 
We presented \gls{mint}, a novel unsupervised keypoint representation learning method from videos using entropy-based intrinsic supervisory signals. We treat keypoints as \textit{transmitters} of information, and defined a deep model that learns consistent keypoint representations from video frames, thanks to two original losses; an information maximization loss and an information transportation loss. These losses drive the keypoints to cover areas of high spatial entropy, while ensuring spatio-temporal keypoint consistency. Auxiliary losses enable \gls{mint} to learn to switch on/off keypoints when required to preserve the information flow.
% , or to switch the redundant ones off when information is contained.
Our experimental evaluation showcased the superior performance of our method on various downstream tasks, ranging from object detection to dynamics prediction and imitation learning. Moreover, we showed qualitatively that \gls{mint} tackles key challenges
% for spatio-temporal keypoint detection 
in realistic scenarios, such as attending to static and dynamic objects and handling appearing/disappearing entities.
Overall, we proposed a method for learning reasonable keypoint representations from videos purely unsupervised, with promising results for future applications.
%Overall, our purely unsupervised learned keypoint representations from videos show promising results for various applications.
%We envision the potential impact of \gls*{mint} for applications like explainable video prediction, robot learning from video demonstrations, and scene understanding.
\section*{Acknowledgements}
This work is funded by the DFG Emmy Noether Programme No. CH 2676/1-1, and the Hessian.AI through the Connectom Fund on Lifelong Explainable Robot Learning. The project has also been supported in part by the State of Hesse through the cluster project “The Third Wave of Artificial Intelligence (3AI). We would like to thank Stefan Roth, Carlo D'Eramo, and An Thai Le for the discussions and comments. We would also like to thank the anonymous ICML reviewers for their valuable feedback on the manuscript.
\clearpage
\bibliography{icml_bibliography}
\bibliographystyle{icml2023}

%%%%%%%%%%%%%%%%%%%%%%%%%%%%%%%%%%%%%%%%%%%%%%%%%%%%%%%%%%%%%%%%%%%%%%%%%%%%%%%
%%%%%%%%%%%%%%%%%%%%%%%%%%%%%%%%%%%%%%%%%%%%%%%%%%%%%%%%%%%%%%%%%%%%%%%%%%%%%%%
% APPENDIX
%%%%%%%%%%%%%%%%%%%%%%%%%%%%%%%%%%%%%%%%%%%%%%%%%%%%%%%%%%%%%%%%%%%%%%%%%%%%%%%
%%%%%%%%%%%%%%%%%%%%%%%%%%%%%%%%%%%%%%%%%%%%%%%%%%%%%%%%%%%%%%%%%%%%%%%%%%%%%%%
\newpage
\appendix
\onecolumn
\section*{Appendix}
The appendix provides additional details on the architecture, the entropy layer, proofs, the description of evaluation metrics for object detection and tracking task, additional experimental analysis with implementation details and discussions, and an extended related work section.
Moreover, we provide video results on the project website\footnote{ \url{https://sites.google.com/view/mint-icml}} and the code\footnote{\url{https://github.com/iROSA-lab/MINT}}.

\startcontents
  \printcontents{}{1}{}

%We include details to help in reproduce the MINT. 
% \pagebreak
% \begin{appendices}
\section{Architecture}
This section contains additional information about the model architecture in~\cref{sec:arch}, implementation details on how to get the keypoint coordinates from the feature map, and the heatmaps for keypoints to ensure reproducibility. 

\subsection{Keypoint Model} 
The backbone of our model is an hourglass architecture ~\citep{hourglass}, followed by a soft-argmax operator to receive the coordinates of the keypoints.
Keypoints provide a low-dimensional representation of high-dimensional RGB images. Therefore, many keypoint detection techniques are inspired by the autoencoder architecture~\citep{deep} which uses the bottleneck to consolidate the information into a reduced dimensionality for keypoint extraction~\citep{videostructre,transporter}.
Instead, we suggest taking an hourglass architecture~\citep{hourglass} which upscales the compressed information again and outputs several feature maps with high activation in places with eminent information~\citep{ewerton,graph, hourglass}.
This allows the network to predict the information at the original image size yielding finer resolution of the coordinates and correspondence to the original pixels.
% The hourglass architecture output several feature maps with high activation in places with eminent information~\citep{ewerton,graph, hourglass}. This allows the network to predict the information at the original image size yielding finer resolution and correspondence to the original pixels. 
% Keypoints provide a low-dimensional representation of high-dimensional RGB images. Therefore, 
% In comparison, many keypoint detection techniques are inspired by the autoencoder architecture~\citep{deep}, and use the bottleneck to consolidate the information into a reduced dimensionality for keypoint extraction~\citep{videostructre,transporter}.

Our keypoint detector consists of an hourglass convolutional neural network with three convolutional layers, with kernel sizes of $5,3,3$ and strides of $3,2,2$, respectively. 
% The input channels of the first layer are the same as the number of channels $C$ of the input image, while the number of output channels is the same number of keypoints $K$. The second layer has $K$ input channels and $2K$ output channels. 
The upsampling part of the model consists of three transposed convolutional layers, with kernel sizes of $3,3,3$ and strides of $1,2,2$. 
% The first layer has input with $2K$ filters and output of $2K$ filters. The last layer has input $2K$ and output with K channels, each of which channel corresponds to a keypoint.
The number of input and output channels for each layer depends on the number of keypoints $K$ and the number of input image channels $C$, see \cref{table:model}.
The result is passed through a softplus layer to ensure the positivity of the feature maps. Lastly, we append a spatial soft-argmax layer (see \cref{sec:spatial}) to get the coordinates of the keypoints from the feature maps $f_i$. We initialize all the convolutional layers with Xavier's normal initialization \citep{glorot2010understanding} and add a leaky ReLU activation and a batch normalization layer after each of them. We normalize the input to the range $[-0.5,0.5]$. The total number of the parameters is $58,725$ for the input of size 320$\times$420. %Table \ref{table:model} shows the details of the model architecture.%
% The keypoint detector is the only model needed during training and inference.
% and its small size makes it suitable for a low-resource setting and deployment on onboard ships.

\begin{table}[h]
    \centering
    \caption{%Architecture details with output sizes for [-1,3,240,320] input and 25 keypoints
Architecture details for an RGB image of $320 \times 480$ and $K=25$ keypoints. There is a leaky ReLU layer and a BatchNorm2d layer (50 parameters) after each convolutional layer.}
\resizebox{\textwidth}{!}{%
\begin{tabular}{lcccccc}
\toprule
Layer (type) & Input channels & Output channels & Kernel size & Stride & Output shape &  \# params           \\
\cmidrule(r){1-7}
Normalize-1 & C & C & - & - & [3, 320, 480] & 0 \\
Conv2d-1   & C & K & 5 & 3 & [25, 79, 106] & 1,900\\
Conv2d-2  & K & K & 3 & 2 &         [50, 26, 35]      &    5,650\\
Conv2d-3  & K & 2K & 3 & 2 &         [50, 26, 35]      &    11,300\\
ConvTranspose2d-1  & 2K & 2K & 3 & 1 &         [50, 53, 71]  &          22,550\\
ConvTranspose2d-2   & 2K & K & 3 & 2 &       [25, 107, 143]   &       11,275\\
ConvTranspose2d-3   & K & K & 3 & 2 &       [25, 107, 143]   &       5,650\\
Softplus-14  & K & K & - & - &        =  &              0\\
SpatialSoftargmaxLayer-15        & K & K & - & - &      [25, 2]          &     0\\
\cmidrule(r){1-7}
Final output & & & & & [25,3] & total: 58,725 \\
\bottomrule
\end{tabular}
}
    \label{table:model}
\end{table}

%\captionof{table}{
%Architecture details with output sizes for [-1,3,240,320] input and 25 keypoints
%Architecture details for an RGB-image of $240 \times 320$ and $K=25$ keypoints
%}

\subsection{Feature Map to Keypoint}\label{sec:spatial}
\begin{figure}[t]
    \centering
    \includegraphics[width=0.9\linewidth]{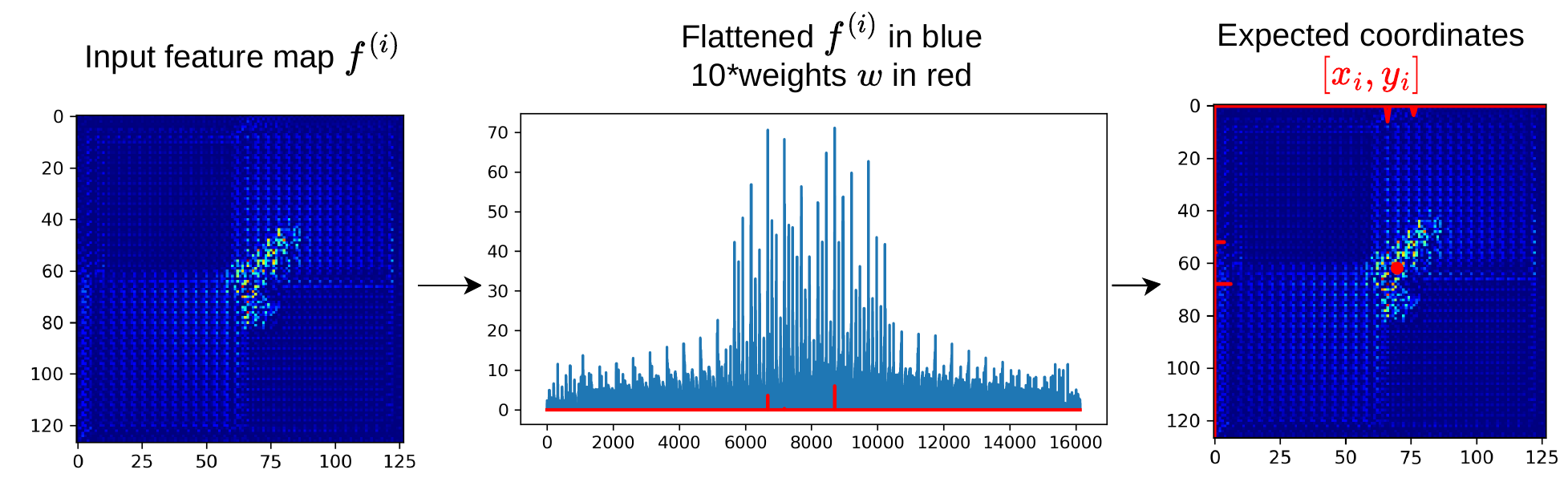}
    \captionof{figure}{Feature map to keypoint (spatial soft-argmax). The keypoint detector outputs a feature map $f^{(i)}$ for each keypoint $k^{(i)}$. The spatial soft argmax operator polls the coordinates of the keypoint $[x_i,,y_i]$  in a differentiable way by flattening a mesh grid of coordinates and using the feature-map values as weights to vote for the coordinates of the maximum value.}
    \label{fig:softmax}
\end{figure}

The coordinates of the keypoints are determined by the location of the maximum value in its corresponding feature map.
The argmax operator is not differentiable, so we opted to use a differentiable spatial soft-argmax as an alternative to extract the keypoints coordinates from the feature maps. The spatial soft-argmax \citep{softargmax} takes $K$ 2D feature maps $f^{(i)}$, one for each keypoint $k_{i}$, flattens the feature maps and computes the weights $\omega_i$ for each pixel
%for the pixels
\begin{equation}
w_i=\frac{e^{f^{(i)}-max(f^{(i)})}}{\sum e^{f^{(i)}-max(f^{(i)})}} \ .    
\end{equation}
%The weights $w_i$ are the softmax of the feature map $f_i$ after applying the subtraction trick. 
Before applying the softmax, we subtract the maximum value from the input, which does not change the output of the softmax but helps for numerical stability.
In order to map the weights to coordinates, we generate
%After generating 
a mesh grid $(x_{grid},y_{grid})$ of $x$ and $y$ coordinates, with the same size as the input image. We flatten the mesh grid and compute the expected keypoint coordinates $[\hat{x}_i, \hat{y}_i]$ as the weighted sum of the coordinate grid with $\omega_i$. This process is visualized in~\cref{fig:softmax}.
%of the soft spatial softmax are the weighted sum of the grid (\cref{fig:softmax}):
% \begin{equation}
% \left\{ \begin{array}{r}
% \hat{x_i}=\sum_{grid} \omega_i*x_{grid} \\
% \hat{y_i}=\sum_{grid} \omega_i*y_{grid} \end{array}
% \right.
% \end{equation}

\begin{figure}[t]
    \centering
    \includegraphics[width=\textwidth]{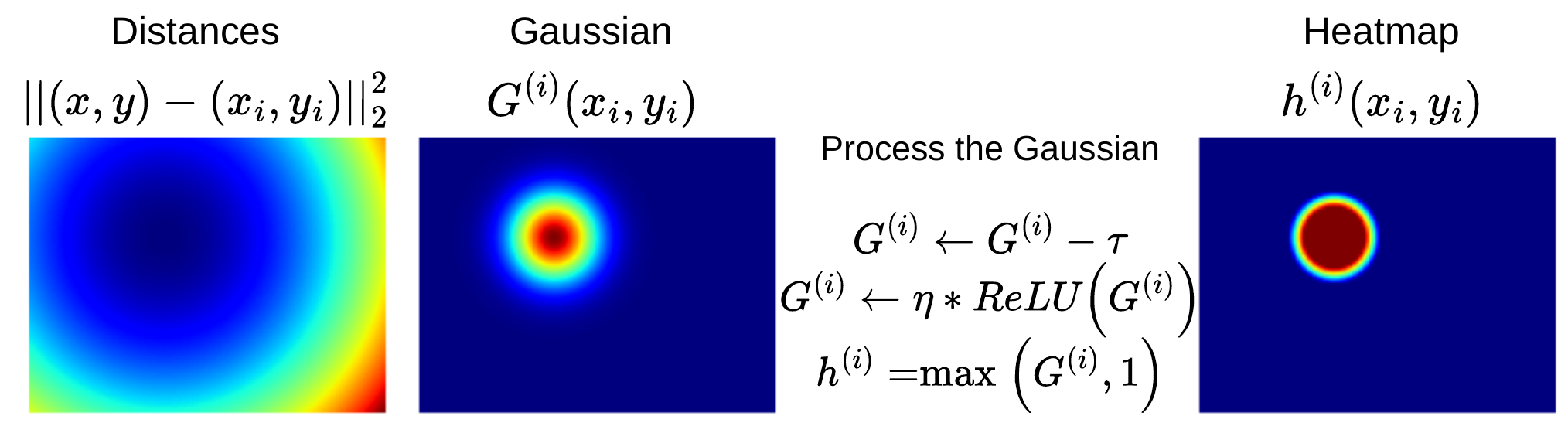}
    \caption{Keypoints to heatmap. We build the heatmap $h^{(i)}$ centered at the coordinates of a keypoint $k^{(i)}$ in a differentiable fashion. The process starts by computing the distances between the center and all the pixels forming a 2D distances image. Inducing the distances into Gaussian forms a multivariate Gaussian distribution  $G^{(i)}$ over the image, whose mean is at the keypoint coordinates. Thresholding and clamping the Gaussian gives the final heatmap, which represents a keypoint's information coverage.}
    \label{fig:heatmap}
\end{figure}

\subsection{Keypoints to Heatmaps}\label{sec:heatmap}
The heatmaps mask out the information coverage areas of keypoints, and are essential to define our losses. We developed a differentiable way to generate heatmaps from keypoint coordinates.
The heatmap $h^{(i)}$ generation for a keypoint takes coordinates as a pair of real numbers $(x_i,y_i) \in \mathbb{R}^2$. We start by generating a pixel-coordinates array with the same width and height as the original image $H\times W\times 2$, where 2 denotes the coordinates of each pixel $(x,y) \in \mathbb{N}^{H \times W \times 2}$.
Then we compute the squared distance between the input and all the pixels $||(x,y)-(x_i,y_i)||^2_2$.
We use the squared distance to generate a Gaussian distribution around the input coordinates $G^{(i)}(x_i,y_i)$ with a standard deviation $\sigma_{G^{(i)}}$. %\footnote{The normalization is omitted in the implementation}
% \begin{equation}
%     G_i(x_i,y_i)=\frac{1}{2\sigma_{G_i}^2 \sqrt{\pi}}e^{-\frac{||(x,y)-(x_i,y_i)||^2_2}{2\sigma_{G_i}^2}}.
% \end{equation}

The heatmap defines the area and weighting of information belonging to each keypoint. The heatmap should be $1$ around the center of the keypoint, as the keypoint covers the information in this point completely, and descend gradually to $0$ representing information out of reach of the keypoint. We achieve that by thresholding and clamping the Gaussian. We use a threshold $\tau$ and a scaling factor $\eta$ for the thresholded Gaussian to get the final heatmap $h_i$. \cref{table:hyperparameters} provides more details on the scale of these hyperparameters. 
% \begin{align}
%     \hat{G_i}(x_i,y_i)&=G_i(x_i,y_i) \\
% \tilde{G_i}(x_i,y_i)&=\lambda \cdot \text{ReLU}(\hat{G_i}(x_i,y_i)) \\
% h_i(x_i,y_i)&=max(\tilde{G_i}(x_i,y_i),1) \\
% \end{align}
The process is visualized in~\cref{fig:heatmap}.

\section{Entropy Layer}\label{sec:a_entropy}
% \todo{Preprocessing, figure + CUDA}
The entropy layer (\cref{sec:entropy}) is one of the main modules of our method. For an input image, the entropy layer outputs the image spatial entropy that is the basis for computing our intrinsic supervisory signals for our representation learning method. In this section, we provide additional implementation details. We split the explanation into two subsections: \emph{(1)} the entropy module definition in PyTorch \citep{pytorch}, and \emph{(2)} the CUDA extension for the parallel execution.

\subsection{Entropy Module}
\begin{figure}[t!]
    \centering
    \includegraphics[width=0.8\textwidth]{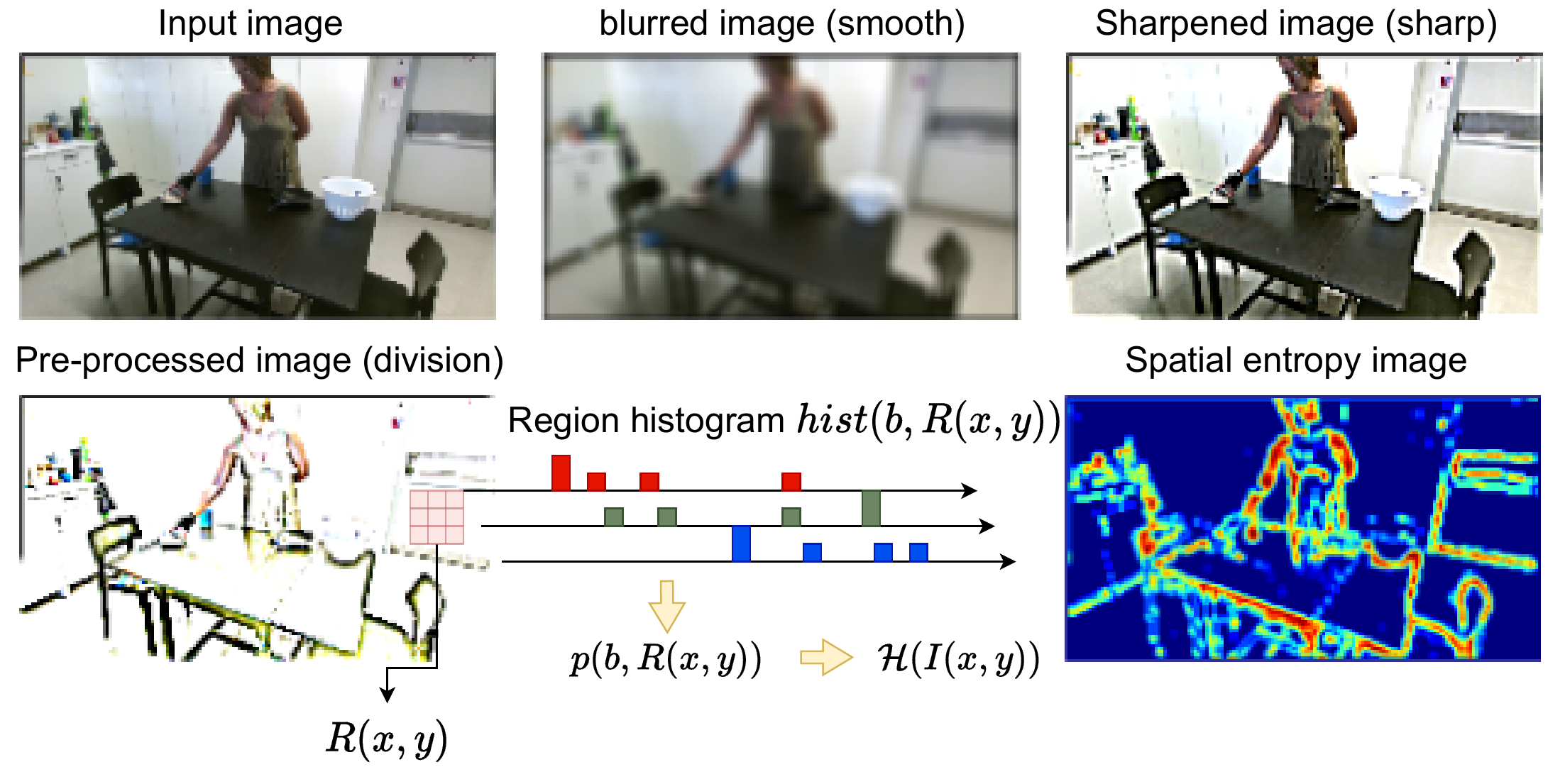}
    \caption{Entropy computation in the entropy layer (\cref{sec:a_entropy}), which consists of the entropy module and the CUDA extension. \textbf{The entropy module} takes as input an RGB image (\textit{input image}), blurs it to get a smoothed image (\textit{smooth} image), and uses the result to sharpen the input image (we get the \textit{sharp} image). The final preprocessed image (\textit{division} image) is the result of dividing the sharpened image (\textit{sharp}) by the blurred image (\textit{smooth}). Further, the entropy module extracts non-overlapping patches and forwards them to the CUDA extension.
    \textbf{CUDA extension} computes the region histograms for each region (patch) $hist(b,R(x,y))$, and uses the histogram to compute the probability of each bin $p(b,R(x,y))$. The entropy of a pixel at location $(x,y)$ is Shannon's entropy of the region around it $\mathcal{H}(I(x,y))$ depending on the probabilities of the color values inside that region. The final \textit{spatial entropy image} is formed by the individual entropies of pixels. Our CUDA extension provides a highly efficient and parallelizable implementation for the process.}
    \label{fig:entropy}
    \vspace{-0.5cm}
\end{figure}

The entropy module is a PyTorch \cite{pytorch} module, which preprocesses an input image and forwards it to the CUDA extension for efficient entropy computation.
The input of the entropy module is an RGB image $I \in \mathbb{R}^{\mathrm{H} \times \mathrm{W} \times 3}$. The input image is first processed to remove high-frequency color changes. The processing is completely vectorized to allow efficient execution using PyTorch during training. The processing of the input image starts by blurring the image using an average blur layer, followed by sharpening the result, and finally dividing the sharp image by the smooth image. \cref{fig:entropy} shows an example of the intermediate steps of the entropy layer.

Before forwarding the processed image to the entropy function, we generate the neighborhood region $R(x,y)$ for each pixel location $(x,y)$. These regions have square shape with the corresponding pixel $(x,y)$ being the center. Instead of iterating over all the pixels with for loops, we use the stridden operation to factorize the extraction of the regions. Using strides to extract the regions aligns with how the image is stored in the memory and does not create overhead. These tricks are essential for the efficiency of our entropy layer.
% \begin{lstlisting}[language=python, caption=Preprocessing and extract patches]
% input=input.view(-1,C,H,W)
% # blur the image
% smooth=self.blur(input).round()
% # sharp the image by add weighted sum of the blurred image
% sharp=torch.clamp(2.5*input-1.25*smooth,min=0,max=255).round()
% # divide the image by the blur
% division=torch.clamp(torch.div(sharp*255,smooth+1e-8),min=0,max=255).round()
% # reshape the input
% input=division.view(N,SF,C,H,W)
% # strides
% sN,sSF,sC,sH,sW=input.stride()
% # instead of iterating using for loops, we can factorize the patches extraction
% # overlapping patches of size (R x R)
% size=(N,SF,C,H-R+1,W-R+1,R,R)
% # strides
% stride=(sN,sSF,sC,sH,sW,sH,sW)
% # generate the patches
% # N x SF x C x H-R+1 x W-R+1 x R x R
% patches=input.as_strided(size, stride)
% patches=patches.contiguous()
% patches=patches.view(N*SF, C,(H-R+1)*(W-R+1),R*R)
% \end{lstlisting}

\subsection{CUDA Extension}
The CUDA extension is a high-efficient program for parallel image spatial entropy \gls{ise} computation, according to the \gls*{mme}~\citep{brink1996using,razlighi2009comparison}, for which we need the histogram of the color values. 
The naive histogram computation via vectorizing the code requires computing a pairwise distance matrix between each pixel with every histogram bin, corresponding to multiplying the number of possible regions by 256. This causes exploding GPU memory requirements (more than 50GB). Motivated by this observation, we present an efficient entropy layer based on kernel density estimation in this work.

To estimate the value for each histogram bin $b$ inside a region $R(x,y)$ (patch) centered on pixel at location $(x,y)$, we use the kernel density estimator
\begin{equation}
\label{eq:f_hat}
    \hat{f}(b,x,y)= \sum_{(x_n,y_n) \in R(x,y)} \mathcal{K}(\frac{I(x_n,y_n)-b}{B})\ ,
\end{equation}
where $I(x_n,y_n)$ is the pixel value at location $(x_n,y_n)$ inside the region $R(x,y)$, and $B$ is the bandwidth, used as a smoothing parameter.\\
We follow \citep{deephist} and use the derivative of the logistic regression function, the Sigmoid function $\sigma(.)$, as a kernel $\mathcal{K}(.)$, that is for a variable $v$
\begin{equation}
    \mathcal{K}(v)=\frac{d}{dv} \sigma(v) = \sigma(v) \sigma(-v) \ .
\end{equation}
The integral of the function $\hat{f}(b,x,y)$ defined in~\cref{eq:f_hat} over the region gives the histogram value of the bin $b$ in a color channel $c$:
\begin{equation}\label{eqn:hist}
hist_c(b,R(x,y))=\sum_{(x_n,y_n) \in R(x,y)} \left[ \sigma (\frac{I_c(x_n,y_n)-b-L/2}{B}) - \sigma (\frac{I_c(x_n,y_n)-b+L/2}{B})\right] \ ,
\end{equation}
where $L=1/256$ is the bin size, so that each bin represents a color value.
We get the probability of each color value by dividing the sum of the histogram values by the size of the region $|R|$ and the number of channels $C$
\begin{equation}\label{eqn:prob}
p(I(x,y))=p(b,R(x,y))= \frac{1}{C\cdot|R|} \sum_{c\in\{r,g,b\}} hist_c(b,R(x,y)) \ .
\end{equation}
The entropy of the pixel in the center of the patch is
\begin{align}
    \mathcal{H}(I(x,y))&= -\sum_{(x_n,y_n)\in R(x,y)}  p(I(x_n,y_n))\log (p(I(x_n,y_n))) \label{eqn:a_entropy1}\\
            &= - \sum_{b\in [0,255]} p(b,R(x,y)) log(p(b,R(x,y))) \ . \label{eqn:a_entropy2}
\end{align}

The entropy module uses our entropy function, implemented as an autograd function in PyTorch, to realize the CUDA extension of the entropy computation. The input to the entropy function are the regions of the preprocessed images $R(x,y)$. The CUDA extension allocates a GPU block for each region, hence, the grid size equals to the number of all possible regions for all images in the batch. The block size is $256$ threads, \ie, a thread for each bin $b$. Each thread iterates over the whole region and computes the histogram of its corresponding bin value $b$ according to \cref{eqn:hist}. Then, it normalizes the result by the region size and the number of channels to get the probability according to \cref{eqn:prob}. Finally, each thread computes the entropy of the pixel (\cref{eqn:a_entropy1}), which is equivalent to the sum of the entropy of the histogram bins (\cref{eqn:a_entropy2}).\footnote{The entropy layer is opensourced in \url{https://github.com/iROSA-lab/MINT}}

\section{Proofs}\label{sec:proofs}

\subsection{Proof of \cref{propos:max_entropy}}\label{app:proof_max_entropy}
We can consider the network as an information channel similar to the information maximization principle (InfoMax) \citep{linsker1988self}. The input of the network is the actual image $I_t$, and the output is the masked image $I_t^{M}$ by the keypoints.
We want to minimize the average probability of error of how well the output $I_t^{M}$ represents the information of the input $I_t$. First, we will work on a pixel level to bound the probability of error in the intensity of the $n^{\text{th}}$ pixel at location $(x,y)$ in images $I_t^M$ and $I_t$, denoted as $P_{\varepsilon}^{(n)} = \mathbb{P}(I_t(x,y) \neq I_t^M(x,y))$. Images can in general be considered as lattices, with pixels being the random variables over intensities $\mathcal{B}$ (in our case these are the number of bins in the histogram as described in \cref{sec:a_entropy}).

Since the error event for the $n^{\text{th}}$ pixel is a binary event, it follows that $P_{\varepsilon}^{(n)}$ is a binary probability. Therefore, the average error probability over all pixels $N$ of the image can be computed as 
\begin{equation}\label{eq:avg_err_prob}
 \bar P_{\epsilon} = \frac{1}{N}\sum_n^{N} P_{\varepsilon}^{(n)} \ ,   
\end{equation}
where $N = H \times W$ is the total number of pixels, computed as the product of the height $H$ and width $W$ of the image.

On a pixel-level, following Fano's inequality\footnote{Fano's inequality uses information-theoretic measures to provide the relation between the average information loss from a noisy channel and the probability of categorization error.}~\citep{sabuncu2006entropy,scarlett2019introductory,tandon2014information} and assuming that the $n^{\text{th}}$ pixel in position $(x,y)$ in an image can take a value uniformly on $\mathcal{B}$, we get
\begin{equation}\label{eq:fanos_1}
    \mathcal{H}(I_t(x,y)| I_t^{M}(x,y)) \leq \mathcal{H}_2(P_{\varepsilon}^{(n)}) + P_{\varepsilon}^{(n)} \log(|\mathcal{B}|-1) \ ,
\end{equation}
where $\mathcal{H}_2(\alpha) = \alpha \log\frac{1}{\alpha} + (1 -\alpha) \log \frac{1}{1-\alpha}$ is the binary entropy function (with maximum entropy corresponding to $\alpha=\frac{1}{2}$), and $\mathcal{B}$ is the support of the pixel value. 

\cref{eq:fanos_1} can be further bounded as $\mathcal{H}_2(P_{\varepsilon}^{(n)})\leq \log2$, and $|\mathcal{B}|-1\leq |\mathcal{B}|$. Moreover, since a pixel is uniform on $|\mathcal{B}|$ its entropy can be considered $\mathcal{H}(I_t(x,y))= \log |\mathcal{B}|$. Therefore, we can further bound \cref{eq:fanos_1} as 
\begin{align}
   & \mathcal{H}(I_t(x,y)| I_t^{M}(x,y)) \leq \log2+ P_{\varepsilon}^{(n)} \log(|\mathcal{B}|) \\
  \Leftrightarrow& \mathcal{H}(I_t(x,y)| I_t^{M}(x,y)) - \mathcal{H}(I_t(x,y)) \leq \log2+ P_{\varepsilon}^{(n)} \log(|\mathcal{B}|) - \mathcal{H}(I_t(x,y))\\
  \Rightarrow  &\mathcal{I}(I_t(x,y), I_t^{M}(x,y)) \geq  (1-P_{\varepsilon}^{(n)}) \log(|\mathcal{B}|) - \log2  \\
\Leftrightarrow & P_{\varepsilon}^{(n)} \geq 1 - \frac{\mathcal{I}(I_t(x,y), I_t^{M}(x,y))+ \log2}{\log |\mathcal{B}|} \ .\label{eq:fanos_err_mi} 
\end{align}
\\
% & P_{\varepsilon}^{(n)} \geq 1 - \frac{\mathcal{H}(I_t(x,y)) + \mathcal{H}(I_t^{M}(x,y))- \mathcal{H}(I_t(x,y), I_t^{M}(x,y))+ \log2}{\log |\mathcal{B}|} \label{eq:fanos_err_mi2}
The mutual information of two random variables is upper bounded by the minimum entropy of the marginals, therefore,  $\mathcal{I}(I_t(x,y), I_t^{M}(x,y)) \leq \min(\mathcal{H}(I_t(x,y)), \mathcal{H}(I_t^{M}(x,y)))$. But the masked image will by definition represent less information than the original image, therefore, we have $\mathcal{I}(I_t(x,y), I_t^{M}(x,y)) \leq \mathcal{H}(I_t^{M}(x,y))$. We can take the \textit{worse case scenario} and assume $\mathcal{I}(I_t(x,y), I_t^{M}(x,y)) \approx \mathcal{H}(I_t^{M}(x,y))$ \citep{murphy2022probabilistic}.  Therefore, \cref{eq:fanos_err_mi} becomes
\begin{align}
    P_{\varepsilon}^{(n)} &\geq 1 - \frac{\mathcal{H}(I_t^{M}(x,y)))+ \log2}{\log |\mathcal{B}|}\ , \label{eq:fanos_final_me}
\end{align}
which bounds the error on the information carried by a pixel in the masked image.

To acquire the bound of the average error probability in \cref{eq:avg_err_prob}, we sum \cref{eq:fanos_final_me} for all pixels and divide by $N$ to get 
\begin{align}
    \frac{1}{N} \sum_n^{N} P_{\varepsilon}^{(n)} &\geq \frac{1}{N} \sum_n^{N} 1 - \frac{\sum_{x,y}\mathcal{H}(I_t^{M}(x,y))+ \sum_n^{N}\log2}{N \log |\mathcal{B}|} \\
    \bar P_{\varepsilon} &\geq 1- \frac{\sum_n^N \mathcal{H}(I_t^{M}(x_n,y_n))}{N \log |\mathcal{B}|} - \frac{\log2}{\log |\mathcal{B}|}\ . 
\end{align}
\qed

% \subsection{Proof of \cref{lemma:combined_me_mce}}
% The probability of jointly minimizing the error of using the masked image and the masked conditional image, can be seen as the joint probabilities of the events $\varepsilon$ and $\varepsilon^{\text{cond}}$. But these events are not independent, hence, $\bar P_{\varepsilon, \varepsilon^{\text{cond}}} \leq \bar P_{\varepsilon} + \bar P_{\varepsilon^{\text{cond}}}$. Adding the results from \cref{eq:bound_me} and \cref{eq:cond_entropy} and 

% For a discrete random variable $X$ the entropy is
% \begin{equation}
%     \mathcal{H}(X)= - \sum_X p(x) log(p(x)) dx
% \end{equation}
% The conditional entropy for the random variable $X$ given another random variable $Y$ is a positive value and equals
% \begin{equation}
%     \mathcal{H}(X|Y)= - \sum_{(X,Y)} p(x,y) log(\frac{p(x,y)}{p(y)}) dx \geq 0
% \end{equation}
% Similarly, $\mathcal{H}(Y|X)\geq 0$.

% Given that the conditional entropy is always positive, this leads to
% \begin{equation}
%     \mathcal{H}(X,Y) \geq max(\mathcal{H}(X),\mathcal{H}(Y))
% \end{equation}
% Accordingly, the joint entropy $\mathcal{H}(X,Y)$ is bounded by the max of the entropies of the two random variables $max(\mathcal{H}(X),\mathcal{H}(Y))$.
\subsection{Proof of \cref{lemma:joint_entropy}}\label{app:proof_joint_entropy}
We first derive the complete proof for the relation $\mathcal{H}(X,Y) \geq \max(\mathcal{H}(X), \mathcal{H}(Y)) \geq 0
$ that holds for any two discrete random variables, as also stated in \citep{murphy2022probabilistic}-Equation(6.10). Then, we extend this proof for the case of joint \gls{ise}.

Let $X,Y$ be two discrete random variables, and the respective entropies are lower-bounded  $\mathcal{H}(X)\geq0$, $\mathcal{H}(Y)\geq0$. The joint entropy between the two random variables can be expressed as
\begin{equation}\label{proof:joint_ent}
    \mathcal{H}(X,Y)=\mathcal{H}(X)+\mathcal{H}(Y|X)=\mathcal{H}(Y)+\mathcal{H}(X|Y) \ .
\end{equation}
As any conditional entropy is greater or equal to zero,
% is less than the entropy of the random variable, 
we get $\mathcal{H}(X,Y)\geq \mathcal{H}(X)$, and similarly $\mathcal{H}(X,Y)\geq \mathcal{H}(Y)$.
If $\mathcal{H}(X)\geq \mathcal{H}(Y)$ then
\begin{equation}\label{proof:joint_ent_1}
    \mathcal{H}(X,Y) \geq \mathcal{H}(X) \geq \mathcal{H}(Y) \ .
\end{equation}
If $\mathcal{H}(Y)\geq \mathcal{H}(X)$ then
\begin{equation}\label{proof:joint_ent_2}
    \mathcal{H}(X,Y) \geq \mathcal{H}(Y) \geq \mathcal{H}(X) \ .
\end{equation}
Therefore, for discrete random variables, we get the lower bound of the joint entropy as
\begin{equation}\label{proof:joint_ent_3}
    \mathcal{H}(X,Y) \geq \max(\mathcal{H}(X), \mathcal{H}(Y)) \geq 0 \ .
\end{equation}

Following the previous derivation, when considering two images $I_1, I_2$ whose pixels are discrete random variables over intensities $\mathcal{B}$, we can lower bound the joint image spatial entropy by the pixel-wise maximum of the two marginal \gls{ise}
\begin{equation}\label{eq:joint_im_ent_bound}
     \mathcal{H}(I_1,I_2)(x,y) \geq \max(\mathcal{H}(I_1(x,y)), \mathcal{H}(I_2(x,y))) \ .
\end{equation}
\qed

Given this lower bound we can approximate the joint \gls{ise} as 
\begin{equation}\label{eq:joint_im_ent_approx}
     \mathcal{H}(I_1,I_2)(x,y) \approx \max(\mathcal{H}(I_1(x,y)), \mathcal{H}(I_2(x,y))) \ .
\end{equation}

\qed

~\textit{\underline{Remark}.} Approximating the joint entropy by the lower bound corresponds to the worst-case scenario for the \gls{mce} loss. Hence, the approximation ensures better information reconstruction according to \cref{corol:cond_entropy}.
Meanwhile, given that the spatial mutual information of two images is upper bounded by the joint entropy, and since the maximization of the mutual information optimizes the keypoint transportation over frames according to \cref{propos:it}, approximating the joint entropy by the lower bound corresponds to a higher probability of reducing the \gls{it} loss.

\subsection{Proof of \cref{propos:it}}\label{app:proof_it}
Due to the process of information transportation of the keypoints, we try to reconstruct the information each keypoint carries. Therefore, we can again leverage Fano's inequality~\citep{scarlett2019introductory}, to provide a lower bound for the average error probability of information transportation per keypoint. 

We formalize our error probability of information transportation of the $K$ keypoints as the per-pixel error event $P_{\varepsilon}^{\text{IT}} = \mathbb{P}(I_t(x,y) \neq R_t^{(i)}(x,y))$, \ie, aggregating the reconstruction from all keypoints. Therefore, from Fano's inequality similar to \cref{eq:fanos_1}, we have the per-keypoint inequality for each keypoint $k^{(i)}$
\begin{equation}
        H(I_t(x,y)| R_t^{(i)}(x,y)) \leq H_2(P_{\varepsilon}^{\text{IT}(i)}) + P_{\varepsilon}^{\text{IT}(i)} \log(|\mathcal{B}|-1) \ .
\end{equation}
Following similar derivation steps as in \cref{app:proof_max_entropy}, we end up in the equivalent version of \cref{eq:fanos_err_mi}
\begin{align}
& \mathcal{I}(I_t(x,y), R_t^{(i)}(x,y)) \geq  (1-P_{\varepsilon}^{\text{IT}(i)}) \log(|\mathcal{B}|) - \log2 \Leftrightarrow \\
& P_{\varepsilon}^{\text{IT}(i)} \geq 1 - \frac{\mathcal{I}(I_t(x,y), R_t^{(i)}(x,y))+ \log2}{\log |\mathcal{B}|} \label{eq:it_error} \ .
\end{align}

Note that we assume the \gls{it} operation per keypoint, independently, and assuming that it is an exclusive event. Therefore, every single transportation is bounded by \cref{eq:it_error}.
\qed

\section{Evaluation Metrics for Object Detection and Tracking}\label{sec:metrics}

Each keypoint should provide a representation of a feature in an object, and keypoints should be distinctive and distributed over the scene. Keypoints assigned to empty spaces are considered unsuccessfully assigned.
To judge the performance of our method, we propose metrics that use the object masks provided by CLEVRER \citep{clevrer} over a set of test videos $V$, each of which is of length $T$.

\subsection{Percentage of the Detected Objects (DOP)} 
We consider an object detected if there is at least one keypoint on its mask $M_{obj}$.  At each time frame, we count the percentage of detected objects with respect to the ground truth (GT) number of objects and average these values over the whole video. We get the final result by averaging the value over all the videos in the test dataset. Better detection corresponds to a higher percentage of detected objects.

\begin{equation}
    M_{DOP} = \frac{1}{V\cdot T} \sum_{v=1}^{V} \sum_{t=1}^{T} \frac{N_{detected}}{N_{GT}} \ ,
\end{equation}
where $N_{detected}$ is the number of detected objects (at least one keypoint lies in the object mask)
\begin{equation}
    N_{detected} = \sum_{{obj}\in O} \left[\sum_{i=1}^K \mathbb{I}((x_i,y_i) \in M_{obj})\right]>0,
\end{equation}

with $N_{GT}$ being the ground truth number of objects and $O$ is the set of all objects in the scene.

\subsection{Percentage of Tracked Objects (TOP)} 
We consider an object tracked if there is at least one keypoint on its mask in the current and the previous timeframe.  At each time frame, we count the percentage of tracked objects with respect to the ground truth (GT) number of objects and average these values over the whole video. We get the final result by averaging the value over all the videos in the test dataset. Better detection corresponds to a higher percentage of tracked objects

\begin{equation}
    M_{TOP} = \frac{1}{V\cdot T} \sum_{v=1}^{V} \sum_{t=1}^{T} \frac{N_{tracked}}{N_{GT}} \ ,
\end{equation}

where $N_{tracked}$ is the number of tracked objects (at least one keypoint lies in the object mask in time frames t and t-1)
\begin{equation}
    N_{tracked} = \sum_{{obj} \in O} \left[\sum_{i=1}^K [\mathbb{I}((x_i,y_i)_{t} \in M_{obj}^{(t)}]\cdot[\mathbb{I}((x_i,y_i)_{t-1} \in M_{obj}^{(t-1)})]\right] > 0 .
\end{equation}

\subsection{Unsuccessful Keypoint Assignment (UAK)} A keypoint is unsuccessfully assigned in a time frame if it does not belong to any object. We average the number of unsuccessful keypoint over the whole video, and then over test videos to get a global value over the testset

\begin{equation}
    M_{UKA} = \frac{1}{V\cdot T} \sum_{v=1}^{V} \sum_{t=1}^{T} N_{uk} \ ,
\end{equation}
where $N_{uk}$ is the number of unsuccessful keypoints (does not belong to the sum of the masks)
\begin{equation}
    N_{uk} = \sum_{i=1}^K \sim \mathbb{I}((x_i,y_i) \notin \sum_{{obj} \in O} M_{obj})  \ .
\end{equation}
A lower unsuccessful keypoint assignment metric $M_{UKA}$ corresponds to better keypoints activation.

\subsection{Redundant Keypoint Assignment (RAK)}
Assigning keypoints to areas already represented by other keypoints signals bad keypoint detection. The RAK metric accounts for the number of keypoints over the area of the object. The number of keypoints on an object mask should be proportional to its area $A_{obj}$. We assume a keypoint can represent some area of pixels $A_k$. If the keypoints cover the object, the RAK metric will have a value of 0, with higher values if more or fewer keypoints were assigned to that object.
% To get the density, we divide the product of $A_k$ with the number of keypoints on an object mask by the total area of the object
% \begin{equation}
%   M_{DAK} = \frac{1}{V\cdot T} \sum_{v=1}^{V} \sum_{t=1}^{T} \prod_{{obj\in O}} \frac{A_k n_{obj}} {A_{obj}}  \ ,
% \end{equation}
\begin{equation}
  M_{RAK} = \frac{1}{V\cdot T \cdot O} \sum_{v=1}^{V} \sum_{t=1}^{T} \sum_{{obj\in O} } \frac{|A_{obj} - A_k n_{obj}|} {A_{obj}}  \ ,
\end{equation}
where $A_k$ is the representation area of a keypoint (e.g. average object areas in the dataset) $A_{obj}$ is the area of the object’s mask and $n_{obj}$ is the number of keypoints assigned to the object

\begin{equation}
    n_{obj} = \sum_{i=1}^K \mathbb{I}((x_i,y_i) \in M_{obj}) \text{\qquad for each \ } {obj} \in O \ .
\end{equation}
The lower the value of $M_{RAK}$, the better, because more efficient, is the distribution of the keypoints.

The metrics collectively judge the efficacy of keypoint detection and tracking methods, where only detected objects can be tracked, so the DOP metric is an upper bound for the TOP metric. The value of the metric RAK will go to one in the case of not detecting any object, but can go higher in case of assigning redundandant keypoints to the same object. Following this observation, we recommend judging the value of the RAK metric jointly with the value of the DOP metric.

\section{Additional Experimental Analysis}

% \todo{Weights, hyperparameters, datasplit}
\subsection{Ablation Study}\label{sec:ablation}
% The ablation study consists of two major ablations \textbf{MINT w/o Reg.} and \textbf{MINT w/o Temp.}, followed by an ablation analysis for all of the losses and regularizations:
Our method for unsupervised keypoint discovery in video streams uses a collection of information-theoretic losses and some regularizers. In the ablation study, we investigate the role of each component and discuss our design choices. In the following, we analyze different design choices like the entropy region size, the conditional entropy in the information transportation loss, and the regularizers. 

\myparagraph{Ablation analysis.}
Using the proposed evaluation metrics, we analyzed several aspects of \gls{mint} on CLEVRER \citep{clevrer}. We report the results in \cref{table:ablation}. \\
Since we compute the local entropy using the probability of the pixel value in its neighborhood region, we investigated the effect of the region size on the performance by varying the region size while using the information maximization (IM) (\ie, \glsfirst{mae} and \glsfirst{mce} from \cref{sec:infomax}) loss alone. The results show that a region of size $5\times 5$ gives the highest values for the \glsfirst{dop} and \glsfirst{top} metrics. We observe also that increasing the region size led to an increase in \glsfirst{uak}, with a decrease in \glsfirst{rak}; we hypothesize this is due to an over-smoothing effect of the bigger region, which leaks some information outside the objects. We noticed, on the other hand, an increase in the order of 30 minutes in the training time (50\% of the training time) of one seed when increasing the region size by 2. Given the marginal improvement and the need for more resources, we adopted a region size of $3\times 3$ for all of our experiments.

We examined the information-theoretic losses without regularization to ablate the additional hyperparameter $\kappa$ which sets the contribution of the conditional entropy in the \glsfirst{it} loss. The results prove that adding conditional information improves the keypoint detection, with $\kappa =0.5$ giving the best results for \gls{dop} and \gls{top} followed by $\kappa =0.9$. The value of \gls{rak} increases with lower $\kappa$, because keypoints seek the same areas of high information to reconstruct as much information as possible, leading to the redundant assignment. The introduction of conditional entropy in the IT loss, as describe in \cref{sec:it}, helps mitigate this behavior by lowering the reconstruction error outside the transportation regions, \ie, the keypoint position in the current and the previous time frame. We highlight two values from this part; with $\kappa =0.5$ we get the best scores for \gls{dop} and \gls{top}, while $\kappa =0.9$ trades off well all of the metrics (we call this model \colorbox{blue!15}{MINT w/o Reg.} - highlighted in {light blue}, that is also referenced in \cref{table:comparison}).

Next, we investigate the regularization terms proposed in our method: (1) the movement loss controlled by the weight $m_d$ in the information transportation loss, (2) the overlapping loss (O), and (3) the active status loss (S). We experimented with all possible combinations of those regularizers. We can observe that the movement regularizer helps decrease the \gls{uak} metric, as this regularizer stabilizes the keypoint movement and constraints the keypoints from jumping into the background. The overlapping loss reduces the \gls{rak} value by almost half (from 3.982 to 2.079), but this comes with a higher \gls{uak}. The status loss reduces the \gls{uak} but comes at the cost of lower \gls{dop} and \gls{top}. Introducing the overlapping and the status loss together allows better overall performance, where the overlapping loss increases the \gls{dop}. 
% Overall, adding all the regularizers is beneficial for our method, like what we show also in \cref{fig:objectsapp,fig:simitateapp,fig:comapp}.\\
% We argue that optimizing the losses collectively delays the convergence of the overall loss, for fair comparison of overall ablations, we keep the number of epochs the same. 
We achieved the best trade-off across all metrics by setting $\kappa=0.9$ while using all the regularizers (highlighted in {light green}). We adopt this option for our method \colorbox{green!15}{\gls*{mint}}, and it proved to outperform the baselines both in the synthetic dataset (quantitatively proved in \cref{table:comparison}, and qualitatively shown in \cref{fig:objects}) and for realistic scenarios (\cref{fig:comeandgo,fig:simitate}).

Finally, we investigated the performance of the losses that work for single images, mainly the \gls{mae} with the regularizers: the active status loss (S) and the overlapping loss (O). This combination of losses does not use any temporal information, hence, it can operate on static images. We train this combination of losses on CLEVRER, operating on single images. We can observe that the model learns to track objects despite being trained on single images only. However, we argue that the structure of our training process, which uses samples from a sequence of images, biases the model towards reducing the movement of the keypoints while attending to features, leading to good tracking performance. We call this ablation \colorbox{red!15}{MINT w/o Temp.}, and we discuss it further later. 

We show that if we have enough knowledge about the environment and we can decide on the suitable number of keypoint (\eg, K=10 keypoints for CLEVRER), then the \gls{im} loss alone is enough to get good performance (last row in \cref{table:ablation}), with low UAK and RAK, as the keypoint assignment is easier.
% We compared the final loss (MINT) \cref{eq:mint_loss} to the information maximization loss alone (IM), using the information maximization and the information transportation losses (IM+IT) and adding the status loss (IM+IT+S). The results show that the IM loss can outperform all the losses in detecting and tracking objects (highest \textbf{DOP} and \textbf{TOP} values), but assigns a lot of redundant and unsuccessful keypoints (worst \textbf{UAK} and \textbf{RAK}). Adding the IT loss improves the tracking (\textbf{TOP} compared to \textbf{DOP}) and reduces the unsuccessful keypoint assignment (lower \textbf{UAK}). Introducing the specialized status loss S does not help by itself, but its interplay with the overlapping loss allows the overall MINT loss to achieve the desired behavior, \ie, an overall trade-off in good detection and tracking performance and resource assignment. 
\begin{figure*}[t!]
    \captionof{table}{Ablation study on MINT losses. We report the statistics of the metric values over 5 seeds. 
IM stands for the information maximization losses (ME + MCE), IT for information transportation, $\kappa$ decides the contribution of the conditional entropy in the IT loss, $m_d$ is the movement regularizer weight in the IT loss, O is the overlapping loss and S is the active status loss.
The ablations picked for \colorbox{blue!15}{MINT w/o reg.}, \colorbox{red!15}{MINT w/o Temp.} and \colorbox{green!15}{MINT} are highlighted with light blue, light red, and light green consequently.
The weight scales used for all the ablations are $\lambda_{ME}=\lambda_{MCE}=100, \lambda_{IT}=20, \lambda_s=10, \lambda_o=30$, and $K=25$. The * near the method's name indicates a longer training time.}
%   The metrics are described in \cref{sec:obj}.}
  \label{table:ablation}
  \small
  \centering
  \adjustbox{max width=\textwidth}{%
  \begin{tabular}{ccccc}
    \toprule
    Method & \textbf{\gls*{dop} $\Uparrow$} & \textbf{\gls*{top} $\Uparrow$} &\textbf{\gls*{uak} $\Downarrow$}& \textbf{\gls*{rak} $\Downarrow$} \\
    \midrule
    IM (3x3) & 0.951 $\pm$ 0.042 & 0.929 $\pm$ 0.048 & \textbf{6.777 $\pm$ 1.369} & 
    3.885 $\pm$ 1.090 \\
    % 13.020 $\pm$ 1.528 \\
    IM (5x5)* & \textbf{0.956 $\pm$ 0.036} & \textbf{0.932 $\pm$ 0.043} & 8.276 $\pm$ 1.428 & 
    3.660 $\pm$ 1.083 \\
    % 11.939 $\pm$ 1.489 \\
    IM (7x7)* & 0.951 $\pm$ 0.041 & 0.926 $\pm$ 0.048 & 9.946 $\pm$ 1.593 & 
    \textbf{3.098 $\pm$ 0.959} \\
    % \textbf{9.921 $\pm$ 1.746} \\
    
    \hline\hline
    IM+IT ($m_d=0$,$\kappa$=0) & 0.917 $\pm$ 0.072 & 0.897 $\pm$ 0.077 & \textbf{3.543 $\pm$ 1.529} & 
    5.096 $\pm$ 1.587\\
    % 16.094 $\pm$ 1.825\\
    % \rowcolor{blue!15} 
    IM+IT ($m_d=0$,$\kappa$=0.5) & \textbf{0.935 $\pm$ 0.058} & \textbf{0.916 $\pm$ 0.063} & 4.754 $\pm$ 1.463 & 
    3.982 $\pm$ 1.226\\
    % 12.706 $\pm$ 1.925\\
    \rowcolor{blue!15} IM+IT ($m_d=0$,$\kappa$=0.9) & 0.918 $\pm$ 0.073 & 0.897 $\pm$ 0.078 & 6.793 $\pm$ 1.956 & 
    2.478 $\pm$ 0.865\\
    % 7.803 $\pm$ 1.918\\
    IM+IT ($m_d=0$,$\kappa$=1) & 0.916 $\pm$ 0.073 & 0.895 $\pm$ 0.078 & 5.645 $\pm$ 1.873 & 
    \textbf{2.336 $\pm$ 0.768}\\
    % \textbf{7.258 $\pm$ 1.455}\\
    
    \hline\hline
    % IM+IT ($m_d=1$,$\kappa$=0) & 0.879 $\pm$ 0.102 & 0.861 $\pm$ 0.106 & 1.768 $\pm$ 0.967 & 
    % % 2.060 $\pm$ 0.792\\
    % 6.002 $\pm$ 1.617 \\
     IM+IT ($m_d=1$) & 0.883 $\pm$ 0.097 & 0.865 $\pm$ 0.102 & 1.665 $\pm$ 0.954 & 
    1.896 $\pm$ 0.706\\
    % 5.454 $\pm$ 1.400\\
    % IM+IT ($m_d=1$,$\kappa$=0.9) & 0.854 $\pm$ 0.114 & 0.836 $\pm$ 0.117 & 2.142 $\pm$ 1.193 & 
    % % \textbf{1.523 $\pm$ 0.590}\\
    % \textbf{4.293 $\pm$ 1.348}\\
    % IM+IT ($m_d=1$,$\kappa$=1) & 0.872 $\pm$ 0.109 & 0.854 $\pm$ 0.112 & \textbf{1.488 $\pm$ 0.918} & 
    % % 1.621 $\pm$ 0.611\\
    % 4.626 $\pm$ 1.360\\
    % \hline\hline
    % IM+IT ($m_d=0$,$\kappa$=0.5)+O ($\beta=$2) & 0.873 $\pm$ 0.092 & 0.844 $\pm$ 0.099 & 12.499 $\pm$ 1.970 & 
    % % \textbf{1.070 $\pm$ 0.353}\\
    % 3.030 $\pm$ 0.977\\
    IM+IT ($m_d=0$)+O & \textbf{0.921 $\pm$ 0.066} & \textbf{0.898 $\pm$ 0.073} & 7.769 $\pm$ 1.880 & 
    2.079 $\pm$ 0.604\\
    % 6.595 $\pm$ 1.310\\
    % IM+IT ($m_d=1$,$\kappa$=0.5)+O ($\beta=$2) & 0.868 $\pm$ 0.105 & 0.847 $\pm$ 0.109 & 6.438 $\pm$ 1.840 & 1.155 $\pm$ 0.419\\
    IM+IT ($m_d=1$)+O & 0.879 $\pm$ 0.102 & 0.861 $\pm$ 0.105 & 2.196 $\pm$ 1.228 & 
    1.705 $\pm$ 0.582\\
    % 4.955 $\pm$ 1.210\\
    % \hline\hline
    % IM+IT ($m_d=0$,$\kappa$=0.9)+O ($\beta=$2) & 0.877 $\pm$ 0.092 & 0.849 $\pm$ 0.099 & 12.071 $\pm$ 2.112 & 1.093 $\pm$ 0.359\\
    % IM+IT ($m_d=0$,$\kappa$=0.9)+O ($\beta=$4) & \textbf{0.913 $\pm$ 0.070} & \textbf{0.889 $\pm$ 0.076} & 7.931 $\pm$ 2.636 & 1.829 $\pm$ 0.578\\
    % IM+IT ($m_d=1$,$\kappa$=0.9)+O ($\beta=$2) & 0.856 $\pm$ 0.112 & 0.836 $\pm$ 0.115 & 4.922 $\pm$ 1.505 & \textbf{1.074 $\pm$ 0.401}\\
    % IM+IT ($m_d=1$,$\kappa$=0.9)+O ($\beta=$4) & 0.871 $\pm$ 0.107 & 0.853 $\pm$ 0.110 & \textbf{2.374 $\pm$ 1.204} & 1.589 $\pm$ 0.577\\
    % \hline\hline
    IM+IT ($m_d=0$)+S & 0.851 $\pm$ 0.114 & 0.830 $\pm$ 0.118 & 1.057 $\pm$ 0.666 & 
    1.159 $\pm$ 0.455\\
    % 3.197 $\pm$ 0.950\\
    IM+IT ($m_d=1$)+S & 0.842 $\pm$ 0.116 & 0.823 $\pm$ 0.119 & 1.060 $\pm$ 0.735 & 
    1.180 $\pm$ 0.475\\
    % 3.147 $\pm$ 1.095\\
    % \hline\hline
    % IM+IT ($m_d=0$,$\kappa$=0.9)+S & 0.823 $\pm$ 0.125 & 0.801 $\pm$ 0.129 & 1.173 $\pm$ 0.725 & 1.046 $\pm$ 0.401\\
    % IM+IT ($m_d=1$,$\kappa$=0.9)+S & \textbf{0.831 $\pm$ 0.132} & \textbf{0.812 $\pm$ 0.135} & \textbf{1.010 $\pm$ 0.648} & \textbf{1.009 $\pm$ 0.395}\\
    % \hline\hline
    IM+IT ($m_d=0$)+S+O& 0.859 $\pm$ 0.112 & 0.840 $\pm$ 0.116 & 1.130 $\pm$ 0.710 & 
    1.324 $\pm$ 0.508\\
    % 3.865 $\pm$ 1.107\\
    % \rowcolor{blue!15} 
    IM+IT ($m_d=1$,$\kappa$=0.5)+S+O& 0.844 $\pm$ 0.120 & 0.826 $\pm$ 0.123 & 1.100 $\pm$ 0.716 & 
    \textbf{1.121 }$\pm$ 0.451\\
    % 3.035 $\pm$ 1.188\\
    % \hline\hline 
    % IM+IT ($m_d=0$,$\kappa$=0.9)+S+O ($\beta=$4)& 0.843 $\pm$ 0.126 & 0.822 $\pm$ 0.130 & 1.151 $\pm$ 0.645 & \textbf{0.978 $\pm$ 0.376}\\
    % % IM+IT ($m_d=1$,$\kappa$=0.9)+S+O ($\beta=$4)& \textbf{0.847 $\pm$ 0.125} & \textbf{0.827 $\pm$ 0.129} & 1.864 $\pm$ 0.876 & 0.999 $\pm$ 0.385\\
    \rowcolor{green!15} IM+IT ($m_d=1$,$\kappa$=0.9)+S+O & 0.855 $\pm$ 0.118 & 0.838 $\pm$ 0.121 & \textbf{0.889 $\pm$ 0.639} & 
        1.123 $\pm$ 0.448\\
    % \textbf{3.022 $\pm$ 1.156}\\
    \hline\hline
    \rowcolor{red!15} ME+S+O & 0.849 $\pm$ 0.115 & 0.826 $\pm$ 0.119 & 0.958 $\pm$ 0.615 & 
    1.142 $\pm$ 0.446 \\
    % IM+IT (+m,$\kappa$=0.9)+S+O & 0.836 $\pm$ 0.123 & 0.814 $\pm$ 0.128 & 0.983 $\pm$ 0.703 & 1.131 $\pm$ 0.462\\
    \hline\hline
    \rowcolor{blue!15} MINT w/o Reg. & \textbf{0.918 $\pm$ 0.073} & \textbf{0.897 $\pm$ 0.078} & 6.793 $\pm$ 1.956 & 
    2.478 $\pm$ 0.865\\
    % 7.803 $\pm$ 1.918\\
    \rowcolor{red!15} MINT w/o Temp. & 0.849 $\pm$ 0.115 & 0.826 $\pm$ 0.119 & 0.958 $\pm$ 0.615 & 
    1.142 $\pm$ 0.446 \\
    % 3.137 $\pm$ 1.034 \\
    % MINT (ours) &  \textbf{0.847 $\pm$ 0.125} & \textbf{0.827 $\pm$ 0.129} & \textbf{1.864 $\pm$ 0.876} & \textbf{0.999 $\pm$ 0.385}\\
    \rowcolor{green!15} MINT (ours) &  0.855 $\pm$ 0.118 & 0.838 $\pm$ 0.121 & \textbf{0.889 $\pm$ 0.639} & 
    \textbf{1.123 $\pm$ 0.448}\\
    % \textbf{3.022 $\pm$ 1.156}\\
     \hline
    K=10\\
    \hline
    IM (3x3) & 0.879 $\pm$ 0.085 & 0.847 $\pm$ 0.095 & 1.662 $\pm$ 0.781 & 1.536 $\pm$ 0.554 \\
    \bottomrule
  \end{tabular}}
\end{figure*}
We further discuss the two major ablations \colorbox{blue!15}{MINT w/o Reg.} and \colorbox{red!15}{MINT w/o Temp.} in detail regarding their performance qualitatively on the realistic datasets.\footnote{Video results for the ablation study: \url{https://sites.google.com/view/mint-kp/ablations}} 

\begin{figure}[t]
        \centering
        \includegraphics[width=\linewidth]{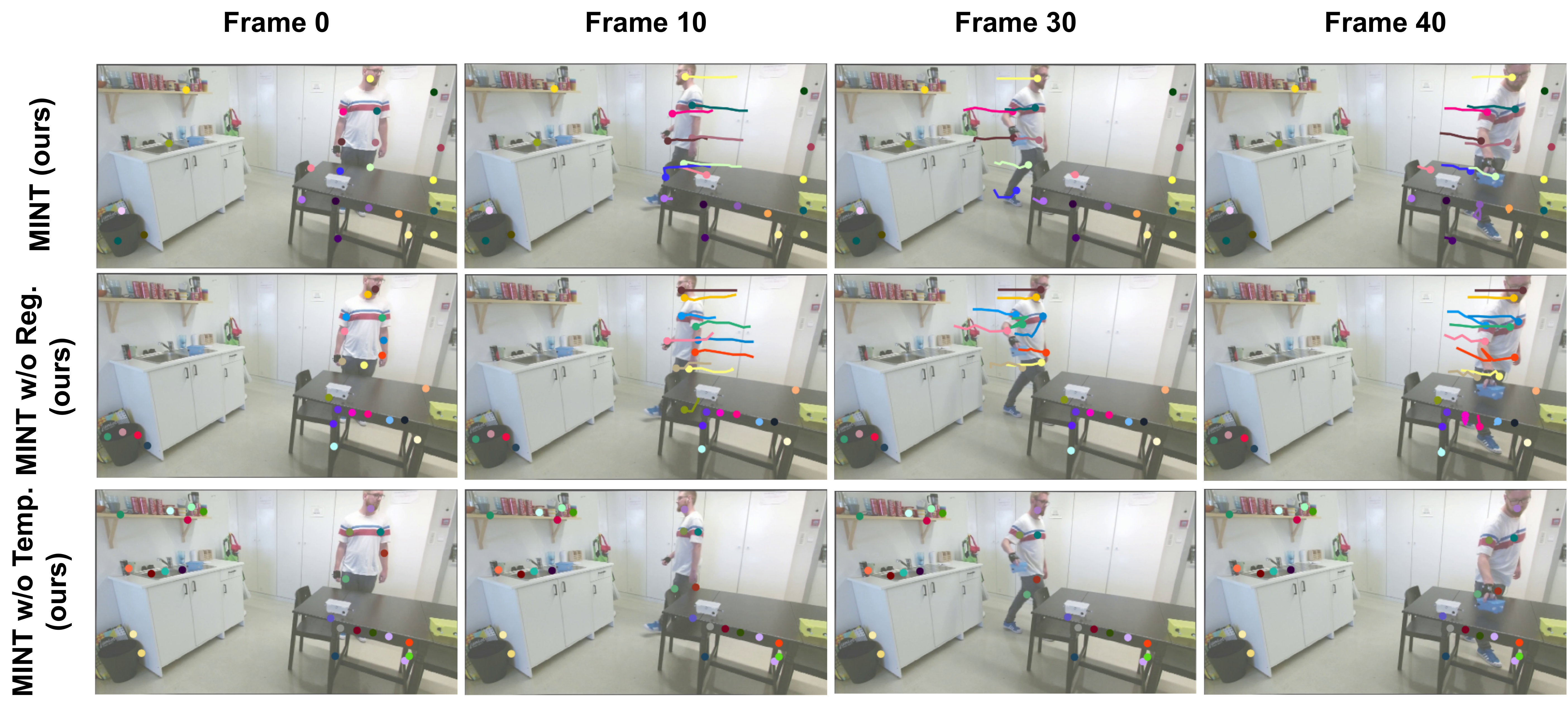}
        \vspace{-0.25cm}
        \caption{Crowded scenes. A video from SIMITATE dataset with a human moving in a room. We compare MINT with its ablations MINT w/o Reg. and MINT w/o Temp. from \cref{sec:ablation}}
        \label{fig:simitateapp}
\end{figure}

\begin{figure}[t]
        \centering
        \includegraphics[width=0.8\linewidth]{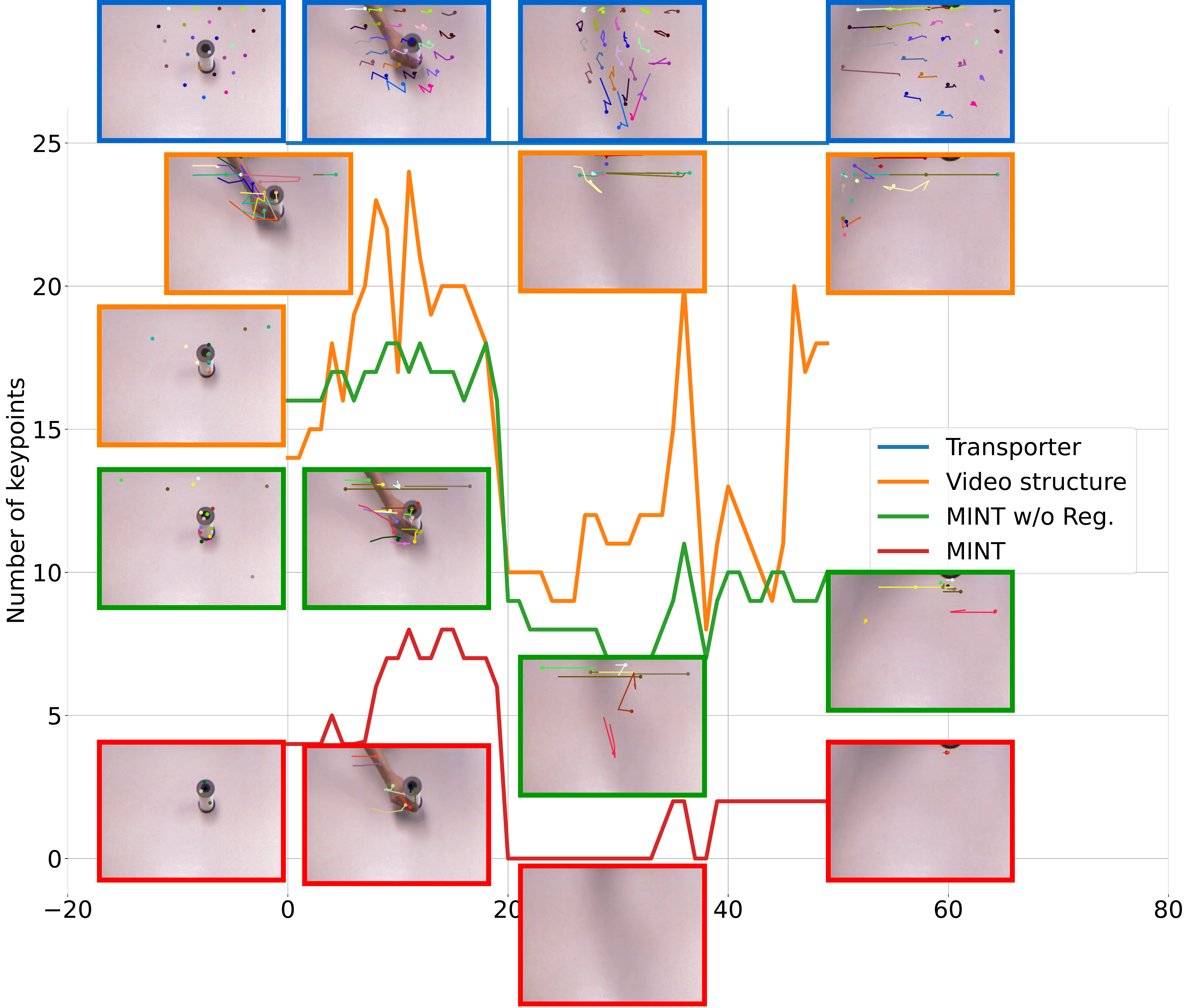}
        \vspace{-0.25cm}
        \caption{Come-and-go scenario. In a manipulation video from MIME, the hand enters after the start of the video and departs before the end. We plot the number of active keypoints w.r.t. timesteps. The results show the role of regularization in our Method \gls{mint} in improving resource assignment and economizing the number of active keypoints.}
        \label{fig:comapp}
\end{figure}

\myparagraph{\colorbox{blue!15}{MINT w/o Reg.}} With MINT w/o Reg. we refer to our method \gls*{mint} without the regularization terms, \ie, (1) removing the regularization for the keypoints' movement $m_d=0$ in the information transportation loss, (2) removing the overlapping loss $\lambda_o=0$, and (3) removing the active status loss $\lambda_s=0$. Besides the quantitative results in \cref{table:ablation}, which show that the information-theoretic losses can detect and track objects better than the baselines (outperforming all of the baselines in the \gls{dop} and \gls{top} metrics), we provide qualitative evidence of the performance of the proposed information-theoretic losses, where MINT w/o Reg. can detect and track the object in synthetic (\cref{fig:objectsapp}) and realistic scenes (\cref{fig:simitateapp,fig:comapp}).

On the other hand, the experiments justify the role of regularization in stabilizing keypoint detection and removing excessive keypoints. \cref{fig:objectsapp} from CLEVRER and \cref{fig:simitateapp} from SIMITATE show a better distribution of keypoint when using MINT over MINT w/o Reg. We refer also to the zoomed-in regions in \cref{fig:objectsapp} where we show MINT w/o Reg. assign keypoints around the object's edges due to the entropy overestimation; MINT regularizes the keypoint towards the center of the object. \cref{fig:comapp} depicts the contribution of the regularization losses for economizing the number of used keypoints in the come-and-go situation, allowing MINT to outperform the other models. 

\myparagraph{\colorbox{red!15}{MINT w/o Temp.}} refers to our method \gls*{mint} operating on single images without the losses that operate temporally over two images. MINT w/o Temp. requires (1) removing the masked conditional entropy loss $\lambda_{\text{MCE}}=0$, and (2) the information transportation loss $\lambda_{\text{IT}}=0$.\\
This ablation proposes good performance for keypoint detection on static images when using the losses that operate on a single image. \cref{table:ablation} shows that MINT w/o Temp. can detect 85\% of the objects in the scene while distributing the keypoint reasonably. \cref{fig:objectsapp} shows that the MINT w/o Temp. assigns keypoints to objects in the scene successfully. \cref{fig:simitateapp} shows that MINT w/o Temp. can detect the human and objects in the background, but due to the lack of temporal information, it does not concentrate on the moving objects (\eg, the hand of the human).

Overall, the full \colorbox{green!15}{MINT} model (\cf\  \cref{table:ablation}) trades off the need for good detection and tracking performance, but with a reasonable distribution of keypoints, to adequately represent the information in the video when minimizing our information theoretic losses (\ie, maximizing the covered information entropy spatio-temporally), as dictated by \cref{propos:max_entropy,propos:it}.

\subsection{Hyperparameters}\label{sec:hyperparameters}
\cref{table:hyperparameters} provides the hyperparameters used for CLEVRER \citep{clevrer} in our experiments. We use the same values for all other datasets, \ie, also for MIME \citep{mime}, SIMITATE \citep{simitate}, and MAGICAL \citep{magical}. The only exceptions are the activation threshold $\gamma$, the std for heatmap $\sigma_{G_i}$ and the threshold of the heatmap $\tau$, where these values depend on the size of the input image (\ie, $\gamma=15$, $\sigma_{G_i}=9.0$, $\tau=0.1$ for MIME, $\gamma=10$, $\sigma_{G_i}=9.0$, $\tau=0.5$ for SIMITATE, $\gamma=10$, $\sigma_{G_i}=7.0$, $\tau=0.3$ for MAGICAL). Our method requires a sequence of 2 frames for the loss computation, and we found that the batch size does not affect the training and can be chosen based on the available GPU resources. In our experiments, we used a PC with a GPU NVIDIA Tesla V100-DGXS-32GB. \gls{mint} consumes around 5GB of GPU memory for a batch size of 32 and trains the model in around 1 hour and 5 minutes (for each seed). We use the same weights for the losses in all experiments over different datasets, which suggests that the model is robust against the hyperparameters. Moreover, we ran additional experiments on our benchmark with different sets of hyperparameters (\cf\ \cref{table:test}), and the results were always close, which provides further proof of the robustness of our method.
% Our method requires at least a sequence of 2 frames for the loss computation.

% Another implementation detail for the information transportation loss, we opt to reduce the contribution of the conditional entropy in building the target entropy image $\mathcal{H}(T_t)=\mathcal{H}(I_t) \odot (h_i(I_t)) + \kappa [\mathcal{H}(I_t|I_{t-1}) \odot (1-h_i(I_t))]$. The conditional entropy is very useful to reconstruct the information, but reducing its contribution gives more importance to track the keypoints.
\begin{table}[t]
\captionof{table}{Hyperparameters}
\label{table:hyperparameters}
\small
\centering
\adjustbox{max width=\textwidth}{%
\begin{tabular}{lc|lc|lc}
\toprule
Parameter name & Value & Parameter name & Value & Parameter name & Value           \\
\cmidrule(r){1-6}
learning rate & 0.001 & clip value & 10.0 & weight decay & 0.00001 \\
epochs & 100 & num keypoints $K$ & 25 &
number of stacked frames & 3 \\ 
activation threshold $\gamma$ & 15 & entropy region size$\sqrt{|R|}$ & 3 &
std for heatmap $\sigma_{G_i}$& 9.0 \\
Threshold for heatmap $\tau$ & 0.1 &
Thresholded heatmap scale $\eta$ & 3.5 & 
CE contribution (IT) $\kappa$ & 0.5 \\
movement weight (IT) $m_d$ & 1.0 &
ME weight $\lambda_{\text{ME}}$ & 100 & MCE weight $\lambda_{\text{MCE}}$ & 100 \\
IT weight $\lambda_{\text{IT}}$ & 20 &
active status weight $\lambda_{\text{s}}$ & 10 & overlapping weight $\lambda_{\text{o}}$ & 30\\
\bottomrule
\end{tabular}}
\end{table}

\begin{figure}
    \captionof{table}{Hyperparameters experiments.}
%   The metrics are described in \cref{sec:obj}.}
  \label{table:test}
  \small
  \centering
  \adjustbox{max width=\textwidth}{%
  \begin{tabular}{ccccc}
    \toprule
    Hyperparams & \textbf{\gls*{dop} $\Uparrow$} & \textbf{\gls*{top} $\Uparrow$} &\textbf{\gls*{uak} $\Downarrow$}& \textbf{\gls*{rak} $\Downarrow$} \\
    \midrule
    $\kappa=0.9$\\
    $\lambda_{IM}=100$\\
    \hline
    $\lambda_{IT}=10, \lambda_s=0, \lambda_o=10, \beta=4$, m=0 & 0.917 $\pm$ 0.070 & 0.894 $\pm$ 0.076 & 7.606 $\pm$ 1.509 & 1.978 $\pm$ 0.593 \\
    $\lambda_{IT}=10, \lambda_s=0.1, \lambda_o=10, \beta=4$, m=0 & 0.913 $\pm$ 0.073 & 0.889 $\pm$ 0.078 & 5.813 $\pm$ 1.325 & 1.887 $\pm$ 0.587 \\
    $\lambda_{IT}=10, \lambda_s=0.1, \lambda_o=10, \beta=4$, m=1 & 0.879 $\pm$ 0.097 & 0.859 $\pm$ 0.102 & 2.468 $\pm$ 1.087 & 1.710 $\pm$ 0.585 \\
    \hline
    $\kappa=0.5$\\
    $\lambda_{IM}=100$\\
    \hline
    $\lambda_{IT}=10, \lambda_s=1, \lambda_o=10, \beta=4$, m=1 & 0.870 $\pm$ 0.107 & 0.852 $\pm$ 0.110 & 2.418 $\pm$ 1.121 & 1.606 $\pm$ 0.574 \\
    $\lambda_{IT}=10, \lambda_s=0, \lambda_o=1, \beta=4$, m=0 & 0.929 $\pm$ 0.062 & 0.907 $\pm$ 0.068 & 8.611 $\pm$ 1.686 & 2.228 $\pm$ 0.696 \\
    $\lambda_{IT}=10, \lambda_s=5, \lambda_o=10, \beta=2$, m=1 & 0.857 $\pm$ 0.115 & 0.838 $\pm$ 0.118 & 1.287 $\pm$ 0.689 & 1.090 $\pm$ 0.438 \\
    $\lambda_{IT}=10, \lambda_s=5, \lambda_o=30, \beta=2$, m=1& 0.851 $\pm$ 0.117 & 0.832 $\pm$ 0.121 & 1.646 $\pm$ 0.706 & 1.050 $\pm$ 0.390 \\
    $\lambda_{IT}=10, \lambda_s=5, \lambda_o=30, \beta=4$, m=1& 0.857 $\pm$ 0.115 & 0.838 $\pm$ 0.118 & 1.102 $\pm$ 0.703 & 1.269 $\pm$ 0.469 \\
    $\lambda_{IT}=20, \lambda_s=5, \lambda_o=30, \beta=4$, m=1& 0.856 $\pm$ 0.117 & 0.838 $\pm$ 0.121 & 1.697 $\pm$ 1.232 & 1.391 $\pm$ 0.544 \\
    $\lambda_{IT}=20, \lambda_s=5, \lambda_o=30, \beta=4$, m=0&  0.859 $\pm$ 0.113 & 0.839 $\pm$ 0.117 & 1.021 $\pm$ 0.599 & 1.283 $\pm$ 0.479\\
    $\lambda_{IT}=20, \lambda_s=5, \lambda_o=30, \beta=2$, m=0&  0.861 $\pm$ 0.107 & 0.839 $\pm$ 0.112 & 1.567 $\pm$ 0.728 & 1.207 $\pm$ 0.462\\
    % $\lambda_{IM}=100, \lambda_{IT}=20, \lambda_s=5, \lambda_o=30, \beta=2, \gamma=10$, m=0&  0.893 $\pm$ 0.080 & 0.865 $\pm$ 0.087 & 9.726 $\pm$ 1.691 & 1.307 $\pm$ 0.391\\
    $\lambda_{IT}=20, \lambda_s=10, \lambda_o=30, \beta=4$, m=1& 0.844 $\pm$ 0.120 & 0.826 $\pm$ 0.123 & 1.100 $\pm$ 0.716 & 1.121 $\pm$ 0.451 \\
    \hline
    $\kappa=0.7$\\
    $\lambda_{IM}=100, \lambda_{IT}=20, \lambda_s=10, \lambda_o=30$\\
    \hline
    $\beta=4$, m=1& 0.845 $\pm$ 0.120 & 0.826 $\pm$ 0.123 & 1.256 $\pm$ 0.975 & 1.041 $\pm$ 0.450 \\
    $\beta=4$, m=0& 0.848 $\pm$ 0.117 & 0.829 $\pm$ 0.121 & 1.270 $\pm$ 0.766 & 1.068 $\pm$ 0.442 \\
    $\beta=2$, m=1& 0.846 $\pm$ 0.120 & 0.826 $\pm$ 0.125 & 1.545 $\pm$ 0.946 & 1.020 $\pm$ 0.384 \\
    $\beta=2$, m=0& 0.839 $\pm$ 0.118 & 0.818 $\pm$ 0.122 & 1.088 $\pm$ 0.647 & 0.948 $\pm$ 0.352 \\
    \hline
    200 epochs\\
    $\lambda_{IM}=100, \lambda_{IT}=20, \lambda_s=10, \lambda_o=30$, m=1\\
    \hline
    $\kappa=0.9, \beta=2$& 0.842 $\pm$ 0.124 & 0.821 $\pm$ 0.128 & 2.121 $\pm$ 0.736 & 1.002 $\pm$ 0.387 \\
    $\kappa=0.5, \beta=4$& 0.835 $\pm$ 0.123 & 0.817 $\pm$ 0.126 & 1.194 $\pm$ 0.797 & 1.047 $\pm$ 0.412 \\
    $\kappa=0.7, \beta=4$& 0.843 $\pm$ 0.126 & 0.825 $\pm$ 0.129 & 1.003 $\pm$ 0.649 & 1.045 $\pm$ 0.419 \\
    $\kappa=0.9, \beta=4$& 0.836 $\pm$ 0.125 & 0.818 $\pm$ 0.128 & 1.441 $\pm$ 1.390 & 1.004 $\pm$ 0.384 \\
    \bottomrule
  \end{tabular}}
\end{figure}

\subsection{Baselines}\label{sec:baselines}

% \textbf{MINT (ours)} our method with all the losses presented in the paper.

\textbf{Video structure} \citep{videostructre} is an unsupervised method for learning keypoint-based representation from videos. Video structure learns a keypoint detector $\phi^{det}(v_t)=x_t$ for a video sequence $v_t$ that captures the spatial structure of the objects in each frame in  a set of keypoints $x_i$. It learns a reconstruction model $\phi^{rec}$ that reconstructs frame $v_t$ from its keypoint representation $x_t$ and the first frame of the sequence $v_1$. An additional skip connection from the first frame to the reconstruction model output changes its actual task to predict $v_t-v_1$; hence $v_t-v_1=\phi^{rec}(v_1,x_t)$.\\
The keypoint detector is trained to optimize three losses:\\
    (1) L2 image reconstruction loss
    \begin{equation}
        \mathcal{L}_\text{image}=\sum_t ||v-\hat{v}||_2^2 \ ,
    \end{equation}
    where $v$ is the true and $\hat{v}$ is the reconstructed image.\\
    (2) Temporal separation loss penalizes the overlap between trajectories within a Gaussian radius $\sigma_{\text{sep}}$
    \begin{equation}
        \mathcal{L}_{\text{sep}}=\sum_k \sum_{k'} exp(-\frac{d_{kk'}}{\sigma_{\text{sep}}}) \ ,
    \end{equation}
    where $d_{kk'}=\frac{1}{T}\sum_t ||(x_{t,k}-\overline{x}_k)-(x_{t,k'}-\overline{x}_{k'})||_2^2$ is the distance between the trajectories of keypoints $k$ and $k'$.\\
    (3) Sparsity loss adds an L1 penalty on the keypoint intensity $\mu$ (the mean value of the corresponding feature map) to encourage keypoints to be sparsely active 
    \begin{equation}
        \mathcal{L}_{\text{sparse}}=\sum_k |\mu_k| \ .
    \end{equation}
    A keypoint is active if the intensity is higher than a specific threshold $th_{\mu}$, the threshold is a hyperparameter that has to be tuned depending on the video.

\begin{figure*}[t]
    \centering
    \includegraphics[width=0.9\linewidth]{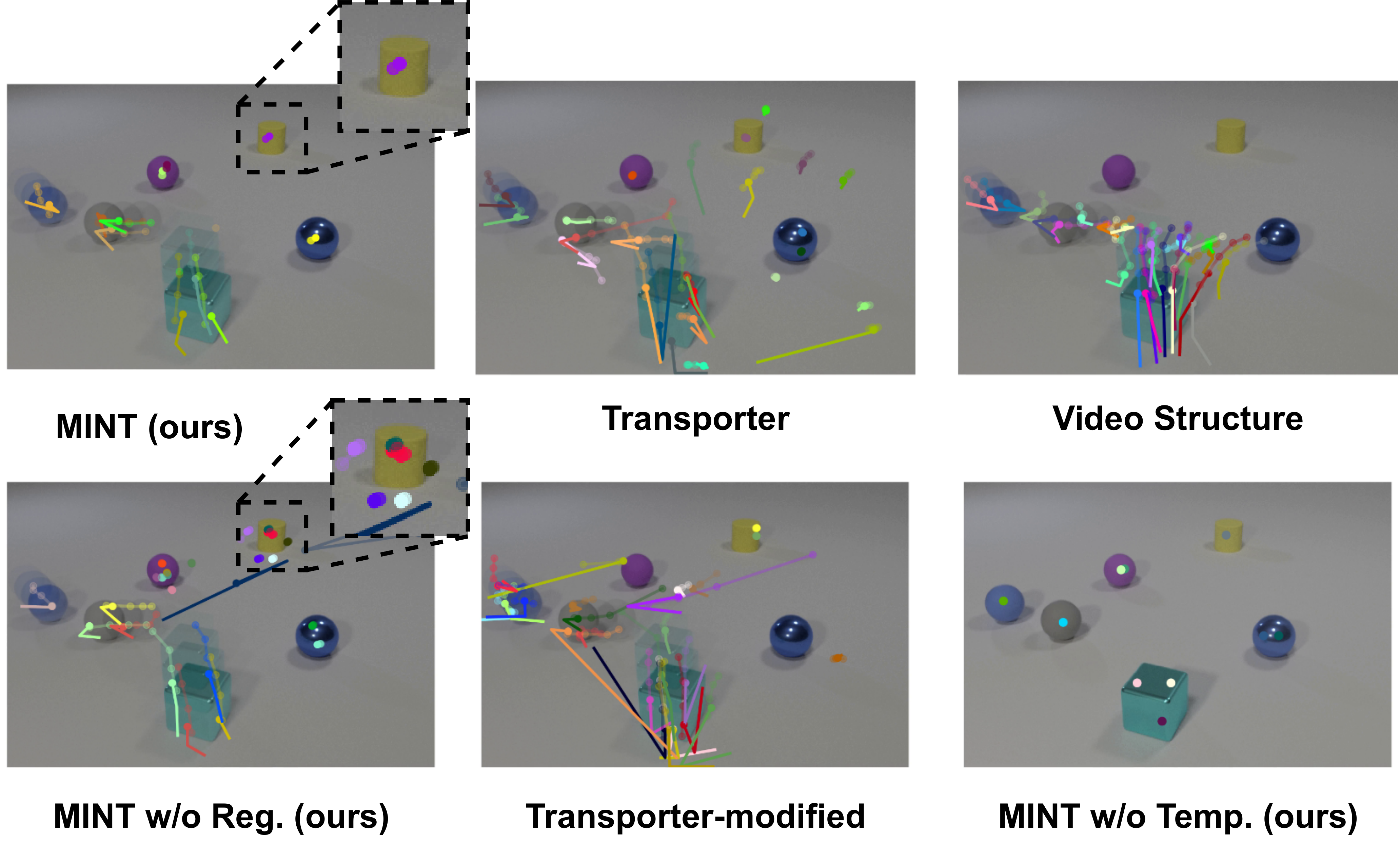}
    % \vspace{-0.7cm}
    \caption{Qualitative results on CLEVRER dataset for Task I (object detection and tracking) and Task II (learning dynamics). We include results for all baselines described in \cref{sec:baselines}.
    }
    \label{fig:objectsapp}
\end{figure*}

\textbf{Transporter} \citep{transporter} is a neural network architecture for discovering keypoint representations in an unsupervised manner by transporting learned image features between video frames using the keypoint bottleneck. During training, spatial feature maps $\phi(x)$ and keypoint coordinates $\psi(x)$ are predicted for a source frame $x_s$ and a target frame $x_t$ using a ConvNet and KeyNet \citep{keynet}. The keypoint coordinates are transformed into Gaussian heatmaps $h_{\psi(x)}$. \\
A transported feature-map $\hat{\phi}(x_s,s_t)$ is generated by suppressing both sets of keypoint location in $\phi(x_s)$ and composing into the feature maps around the keypoints from $x_t$:
\begin{equation}
    \hat{\phi}(x_s,s_t) \triangleq (1-h_{\psi(x)_t})\cdot(1-h_{\psi(x)_t})\cdot \phi(x_s) + h_{\psi(x_t)}\cdot \phi(x_t) \ .
\end{equation}
An additional refiner net learns to map the transported features maps into an image $\hat{x}_t$. The learning objective is reconstructing the target image $x_t$ from the process. Hence, the Transporter optimizes the L2 reconstruction error $\mathcal{L}=||x_t-\hat{x}_t||_2^2$.

\textbf{Transporter-modified} is a modified version of the transporter baseline \citep{transporter}. The original implementation of the method has two potential bottlenecks: (1) the feature maps $\phi(s)$ have a receptive field of size 24 for each position, for an input of size 128x128; and (2) the resolution of the feature maps between which the features are transported is 32x32. For fair comparison to our method, which uses an entropy region of size 3x3, we modified the network architecture of ConvNet $\phi(x)$ to have (1) a receptive field of 7 and (2) a feature map of size 122x122. We call the new architecture Transporter-modified.\\
The experimental results show that the Transporter-modified model outperforms the original Transporter in the quantitative evaluation on CLEVRER dataset \citep{clevrer} (\cf\ \cref{table:comparison}). We want to refer to the visual results in \cref{fig:objectsapp} that show that the keypoints detected by the original Transporter (top middle image) are more stable than those detected by the Transporter-modified (bottom middle image). We argue this behavior is due to the smaller receptive field leading the model to assign keypoints to features instead of objects, and thus keypoints jump to similar features in different objects.

\subsection{Interaction Network Architecture} \label{a_interact}

The interaction network (IN) \citep{battaglia2016interaction} is a model developed for learning the interaction relations between physical objects to infer the physics of the environment. The interaction network treats the objects as nodes of a graph, with the relations as edges. In our case, we use the keypoints as object nodes, with the coordinates, status, and positional encoding as features. We form a fully connected graph of the keypoints, with no relational features for edges.

The interaction network used in our experiments has two sub-models; a relational model and an object model. The relational model uses the relational information and object attributes to predict the effects of all interactions. The object model uses the effects to update the features of the object. We encode node features before passing them to the interaction network. After one pass through the interaction network, we decode the features into coordinates for the prediction task, and we add another prediction head for the action decoding in the imitation learning task.

\subsection{Imitation Learning Results} \label{sec:a_magical}
\textbf{CNN-agent.} The CNN agent is trained from scratch for every environment.
Note that the state space for the CNN agent, the image pixels, is one order of magnitude higher than the keypoints' features. For fair comparison, we train the CNN agent longer (twice the epochs used for training the \gls*{mint}-based agent to counteract for \gls*{mint}'s pretraining).

The CNN feature extractor consists of 5 convolution blocks, each consisting of a 2D convolutional layer with a ReLU activation function and a batch normalization layer. The input to the model is a sequence of 4 color images stacked over the channel axis; hence, the input size is $12 \times 96 \times 96$. The layers have 64, 128, 128, 128, and 128 filters, with a kernel size of 3 and stride of 2, except the initial layer, which has a kernel size of 5 and stride of 1. The output of the last block is flattened and passed to a linear layer to provide the final features. A policy model uses the features to infer the actions. The policy model is a multi-layer perceptron with 4 linear layers of sizes 128, 64, 32, and 32. The output matches the action dimension of MAGICAL environment which is 18.

\begin{figure}[t]
    \centering
    \includegraphics[width=\textwidth]{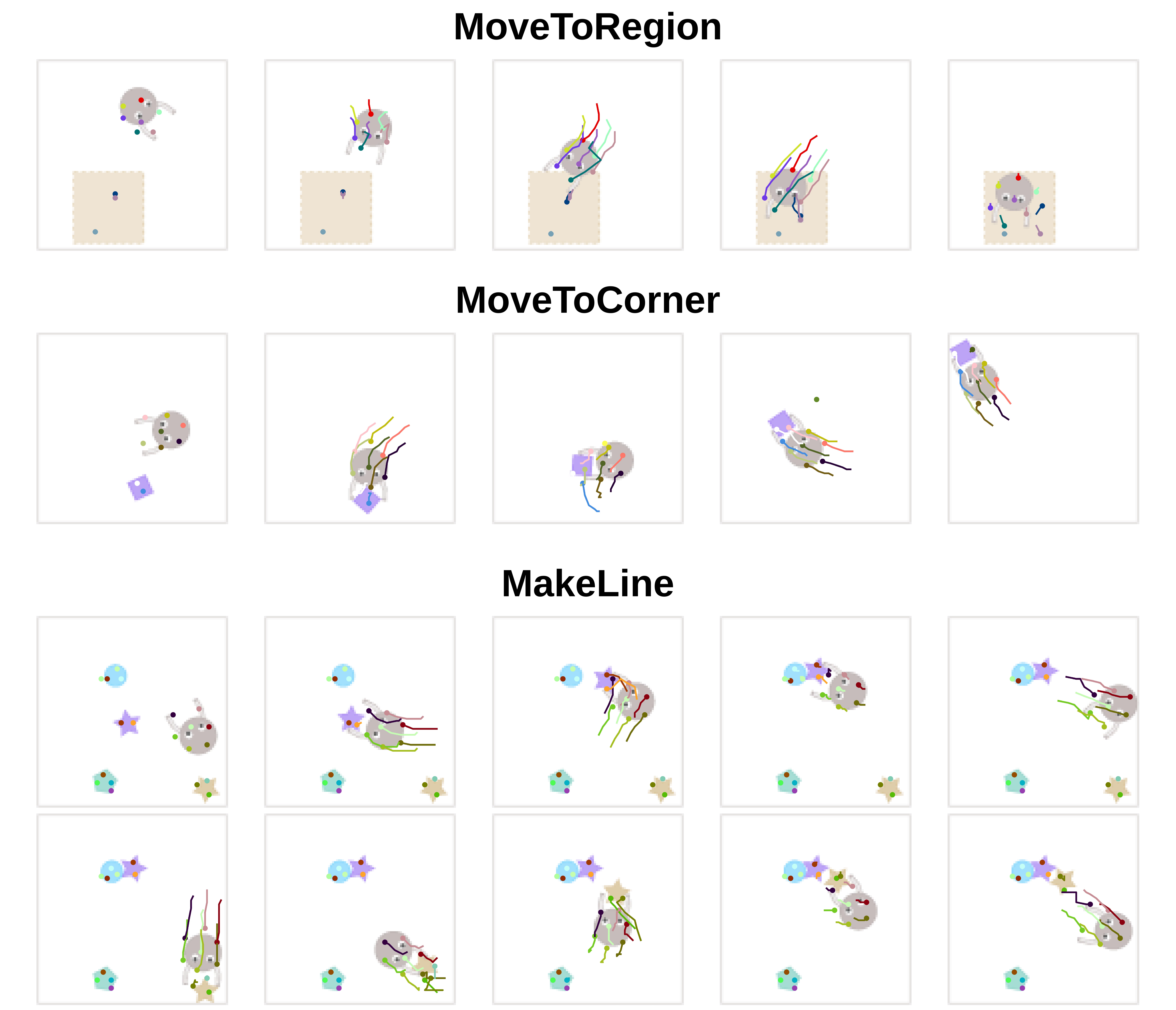}
    \captionof{figure}{Rollouts from MINT agent in MAGICAL \citep{magical} dataset.}
    \label{fig:imitation}
\end{figure}

We also provide visualizations of the learned policies of the \gls*{mint}-based agent on the three environments of MAGICAL in \cref{fig:imitation}.
The visualization shows that MINT can assign reasonable keypoints for the agent and all the objects in the environment. The imitation agent can solve the first two tasks \textbf{MoveToRegion} and \textbf{MoveToCorner}, but it struggles with the last task \textbf{MakeLine}. The agent receives a score of 1.0 when it sorts all 4 objects in one line, while it gets a score of 0.5 for putting 3 out of 4 in one line. Our imitation agent could (occasionally) sort only 3 out of 4 in the depicted environment (which led to an overall 0.2 mean score over 5 seeds -- \cref{table:imitation}), despite being able to assign keypoints to all objects. The results suggest that there is a problem in encoding the relational features between keypoints, hindering the agent to reason upon getting the right locations. We argue that further investigation of the appropriate model to pool information from the keypoints is necessary to solve this most challenging task, but this is out of scope of the current work.

\subsection{Additional Video Results}
We provide additional video results on the website of our project:
\url{https://sites.google.com/view/mint-kp}.

\section{Code}
Our code is available under an open-source license at: \url{https://github.com/iROSA-lab/MINT} . \\
We provide instructions to run the code, with sample datasets to reproduce the results in the paper.

The implementation of video structure \citep{videostructre}\footnote{\href{https://github.com/google-research/google-research/tree/master/video\_structure}{https://github.com/google-research/google-research/tree/master/video\_structure}} uses outdated libraries. Due to compatibility reasons, we reimplemented their code in PyTorch with our best effort. We adapted the implementation of Transporter from \citet{li2020causal} \footnote{ \href{https://github.com/pairlab/v-cdn}{https://github.com/pairlab/v-cdn}} into the codebase of MINT.\footnote{Our baselines implementation is available in our codebase \url{https://github.com/iROSA-lab/MINT}}

\section{Additional Related Work Discussion}\label{sec:extended_related}

\noindent\textbf{Information-theoretic approaches in machine learning.} Information-theoretic principles proved advantageous in training and understanding machine learning models \citep{yu2021information}. Different information measures aim to describe a random variable's behavior due to a probability density function. The probability density function is normally unknown, and machine learning methods usually estimate it \citep{pardo2018statistical, deephist}. In our method, we use kernel density estimation (KDE) \citep{parzen1962estimation} to estimate the probability density function for a region of pixels.
Various information-theoretic quantities were used in machine learning for different applications; examples are the cross-entropy loss for classification \citep{good1992rational, deep}, maximum entropy regularization in reinforcement learning \citep{peters2010relative, haarnoja2018soft}, mutual information for self-supervised learning and interpretability \citep{rakelly2021mutual, zhang2018interpretable}, and KL divergence for training deep energy models \citep{yu2020training}. Our approach uses Shannon's definition of entropy \citep{shannon2001mathematical} to compute the local image entropy. With image entropy, we estimate joint entropy, conditional entropy, and mutual information and develop our information-theoretic losses.\\
% Our proposed method uses low-level entropy for a single image during training, while our temporal losses encourage the model to attend to temporal information change.  
Temporal information plays an important role for many downstream tasks. Recent methods in neural video compression \citep{li2022hybrid} propose to estimate spatial-temporal intra-frame entropy over quantized latent representation. In our work, we opted to use inter-frame entropy estimation with temporal losses that encourage the model to attend to temporal information changes, which proved to provide a strong inductive bias for keypoint detection.
\\
% The information bottleneck is an information-theoretic framework that analyzes learning dynamics in deep neural networks \citep{tishby2000information}. 
The information maximization principle (InfoMax) \citep{linsker1988self} treats the neural network as an information channel and aims to maximize the information transferred through the network. Recent methods in computer vision and natural language processing use the InfoMax principle for self-supervised learning \citep{oord2018representation, hjelm2018learning, kong2019mutual}. Our method adopts the treatment of the neural network as an information channel in the information maximization loss and extends it to treat the keypoints as transmitters of information, while being completely unsupervised.
% The information bottleneck principle helped identify two phases of learning (1) the fitting phase and (2) the compression phase using the mutual information between the input, latent, and output \cite{shwartz2017opening, yu2020understanding}. The interplay between the IM and the auxiliary losses in our method shows a similar behavior, where the IM aims for fitting, then the other losses compress the information. The different ablations of our method in \cref{table:ablation} demonstrate this behavior.
\setlength{\glsdescwidth}{0.8\hsize}
\printnoidxglossary[nogroupskip, style = long]
%%%%%%%%%%%%%%%%%%%%%%%%%%%%%%%%%%%%%%%%%%%%%%%%%%%%%%%%%%%%%%%%%%%%%%%%%%%%%%%
%%%%%%%%%%%%%%%%%%%%%%%%%%%%%%%%%%%%%%%%%%%%%%%%%%%%%%%%%%%%%%%%%%%%%%%%%%%%%%%

\end{document}